\documentclass[letterpaper]{article} 
\usepackage{aaai24}  
\usepackage{times}  
\usepackage{helvet}  
\usepackage{courier}  
\usepackage[hyphens]{url}  
\usepackage{graphicx} 
\usepackage{natbib}  
\usepackage{caption} 
\usepackage{algorithm}
\usepackage{algorithmic}
\usepackage{amsmath}

\usepackage{newfloat}
\usepackage{listings}
\DeclareCaptionStyle{ruled}{labelfont=normalfont,labelsep=colon,strut=off} 
\floatstyle{ruled}
\newfloat{listing}{tb}{lst}{}
\floatname{listing}{Listing}
\title{PPIDSG: A Privacy-Preserving Image Distribution Sharing Scheme with GAN in Federated Learning}
\author {
    Yuting Ma\textsuperscript{\rm 1},
    Yuanzhi Yao\textsuperscript{\rm 2}\thanks{Corresponding authors.},
    Xiaohua Xu\textsuperscript{\rm 1}\footnotemark[1]
}
\affiliations {
    \textsuperscript{\rm 1}University of Science and Technology of China\\
    \textsuperscript{\rm 2}Hefei University of Technology\\
    ytma@mail.ustc.edu.cn, yaoyz@hfut.edu.cn, xiaohuaxu@ustc.edu.cn
}

\begin{document}

\maketitle

\begin{abstract}
Federated learning (FL) has attracted growing attention since it allows for privacy-preserving collaborative training on decentralized clients without explicitly uploading sensitive data to the central server.
However, recent works have revealed that it still has the risk of exposing private data to adversaries.
In this paper, we conduct reconstruction attacks and enhance inference attacks on various datasets to better understand that sharing trained classification model parameters to a central server is the main problem of privacy leakage in FL.
To tackle this problem, a privacy-preserving image distribution sharing scheme with GAN (PPIDSG) is proposed, which consists of a block scrambling-based encryption algorithm, an image distribution sharing method, and local classification training.
Specifically, our method can capture the distribution of a target image domain which is transformed by the block encryption algorithm, and upload generator parameters to avoid classifier sharing with negligible influence on model performance.
Furthermore, we apply a feature extractor to motivate model utility and train it separately from the classifier.
The extensive experimental results and security analyses demonstrate the superiority of our proposed scheme compared to other state-of-the-art defense methods.
The code is available at https://github.com/ytingma/PPIDSG.
\end{abstract}

\section{Introduction}

Federated learning \cite{FedAVG}, which enables clients to train their data locally and upload only model parameters to the server for model aggregation, undoubtedly plays a significant role in autonomous driving \cite{FedAuto}, health care \cite{FL-MRCM,FedDG,HarmoFL}, and other industries in recent years.
However, recent researches \cite{label_inference,label_inference_new,PAFL,SME} demonstrate that adversaries can utilize the parameter information uploaded by clients to carry out attacks, resulting in serious privacy leakage problems.

Several strategies have been proposed to enhance security in the face of privacy threats, such as homomorphic encryption \cite{Homomorphic,HHE} can use an encryption algorithm that satisfies the homomorphic operation property of ciphertext to encrypt shared model parameters and differential privacy \cite{diff,PPGenCDR} can prevent privacy leakage by introducing random noise.
Gradient perturbation \cite{Soteria} aims to perturb data representation to guarantee privacy.
However, the aforementioned methods of protecting sensitive data in FL significantly increase computational overhead or sacrifice efficiency to safeguard sensitive data.
Another possible way is to encrypt training data \cite{EtC,InstaHide} without greatly affecting model accuracy.
These solutions mentioned above all depend on sharing classifier parameters to fulfill the model aggregation against image reconstruction attacks in federated learning.
However, in addition to image reconstruction attacks, clients are vulnerable to membership inference attacks and label inference attacks.
These defenses ignore inference attacks \cite{MIA,label_inference} in federated learning and cannot provide adequate security.

To further show the problem of privacy-preserving methods in FL, we first begin with a theoretical analysis of label inference attacks and then apply these defenses to perform attacks on a variety of datasets.
Because FL mitigates overfitting after model aggregation, we next propose an upgraded membership inference attack that integrates two shadow models instead of a single model considering the server attacker can access model parameters as well as structures.
We also experimentally investigate the feasibility of image reconstruction attacks in FL after protection by defense methods.
Extensive experiments and theoretical studies reveal valuable observations into the relationship between privacy leakage and trained classification model parameter sharing.

Motivated by these observations, we design PPIDSG, a scheme that can defend against privacy leakage while maintaining model utility.
Specifically, 
1) \emph{How to realize privacy protection while simultaneously resisting attacks?}
We leverage a block scrambling-based encryption algorithm to transform the original image domain into the target domain while training the classifier locally;
2) \emph{How to enable federated learning without uploading the classifier parameters?}
We utilize a GAN to capture the target image distribution and upload the generator parameters to share it;
3) \emph{How to maintain classification accuracy?} 
We add an additional classification loss to our generator and introduce an independently trained autoencoder to extract interesting features.
To summarize, this paper makes the following contributions:

\begin{itemize}
\item We present an enhanced membership inference attack in FL by reconstructing classifiers of a victim and others through uploaded model parameters.
\item Theoretical analyses and practical experiments validate that parameter sharing of trained classifiers leads to privacy leakage in federated learning.
\item To the best of our knowledge, a framework combining a GAN and a feature extractor without uploading a trained classifier is first proposed, achieving the balance between privacy-preserving and model utility in FL.
\item Extensive experiments on four datasets compare our scheme with other defenses and results manifest our approach provides a considerably more secure guarantee without compromising accuracy.
\end{itemize}

\section{Related Work}

\subsection{GAN in Privacy Applications}

GAN \cite{GAN} was first proposed in 2014.
The adversarial loss between a discriminator and a generator, which attempts to generate images that are indistinguishable from the real, is GAN's successful secret. 
CycleGAN \cite{cycleGAN} proposed a cycle consistency loss to complete the image-to-image translation with unpaired images.
Owing to GAN's advantage in visual translation, more researchers are considering employing it for defense or attack.
A real-time GAN-based learning procedure \cite{GANattack} was proposed that enables the adversary to generate samples from the target distribution.
DaST \cite{DaST} utilized GAN to train a substitute model and launched a model extraction attack.
DeepEDN \cite{DeepEDN} suggested a medical image encryption and decryption network based on CycleGAN.
FedCG \cite{FedCG} leveraged a conditional GAN to achieve privacy protection against image reconstruction attacks.

\subsection{Attack in Federated Learning}

Traditional attacks in FL include membership inference attacks, property inference attacks, and image reconstruction attacks.
MIA \cite{MIA} raised the membership inference attack: given the black-box access to a model, determine whether a given data record is included in the target dataset.
To build an inference attack model, the adversary needs to create shadow training data for shadow models.
ML-Leaks \cite{ML-Leaks} relaxed the assumption and extended it to more scenarios.
Later, additional technologies \cite{LOGAN,Melis,miadiff} were employed to raise the attack, such as GAN.

Property inference attacks infer particular attributes that hold only for a subset of training data and not for others.
The attribute can be replaced with labels \cite{label_inference_new}.
Image reconstruction attacks exploit gradients that users submitted to the server to restore original samples.
With the guidance of the gradient difference produced by original images and dummy images, DLG \cite{DLG} carried out the minimization optimization.
The extraction of ground-truth labels is first proposed in iDLG \cite{iDLG} as an approach to strengthen the attack.
In GradInversion \cite{GradInversion}, a label recovery algorithm for data in larger batches and a group consistency regularization were utilized to rebuild complex images with high fidelity. 
Then, a zero-shot approach \cite{label_inference} promoted image reconstruction to distributed learning and restored labels even when a batch of labels has duplicate labels.


\section{Privacy Leakage in FL}

\subsection{Attack Setup}

In this paper, we consider a common attack scenario where the attacker is an honest-but-curious server.
It indicates that the attacker adheres to the federated learning protocol without corrupting the training process.
Clients upload their local parameters (gradients or weights) to the server.
These two parameters can be regarded as comparable when all users train only one local epoch between two global aggregations with their all training data.
In this assumption, the attack process is equivalent to a white-box attack, where the attacker acquires the knowledge of model parameters and structures.

\subsection{Label Inference Attack}
\label{LIA}

Each client $C_k$ has a local dataset $\mathcal D_k$ = $\{(x_{i}, y_{i})\}_{i = 1}^{n_k}$, where each sample $(x_{i}, y_{i})$ has a data sample $x_{i}$ and a ground truth label $y_{i}$, and they select $\emph{bs}$ (batch size) of their local datasets for training.
Since most classifiers categorize via the cross-entropy loss function, we define the gradients of loss function w.r.t. network weights $\mathcal{W}$ as:
\begin{equation}
{\nabla_{\mathcal{W}}}{\mathcal{L}}(\textbf{x},\textbf{y}) = - \frac{1}{{bs}}\sum\limits_{i = 1}^{bs} {\sum\limits_{j = 1}^{{n_c}} {{\nabla_{\mathcal{W}}}[{y_i}(j)} } \log y'_i(j)],
\end{equation}
where $y'_i$ is the logit output of the last layer after softmax and $n_c$ is the number of label categories.
When the index $j$ of output is equal to the ground truth, $y_{i}(j) = 1$, else $y_{i}(j) = 0$.
Thus, our goal is to measure the number of images $\sum\nolimits_{i = 1}^{bs} {{y_i}(j)}$ in each label category $j$.
For example, if we observe that $\sum\nolimits_{i = 1}^{bs} {{y_i}(0)}=2$ and $\sum\nolimits_{i = 1}^{bs} {{y_i}(1)}=2$ in the MNIST dataset which has a batch size of four, then we conjecture that there are two labels ``0'' and two labels ``1''.

According to GradInversion \cite{GradInversion}, the gradient of each $x_i$ w.r.t. the network output $z_i$ at index $j$ is ${\nabla _{z_i(j)}}{\mathcal{L}}(x_i,y_i)=y'_i(j)-y_i(j)$.
To achieve our goal from uploading gradients, we define $\mathcal{W}_{m,j}$ as the weights of $m^{th}$ unit of the last hidden layer to the output layer at $j$ index and thus obtain the following equation by using a chain rule:
\begin{equation}
\begin{split}
\sum\limits_{m = 1}^{n_m}{\nabla _{{{\mathcal{W}}_{{{m,j}}}}}}{\mathcal{L}}(\textbf{x},\textbf{y}) &= \frac{1}{{bs}}\sum\limits_{i = 1}^{bs} \sum\limits_{m = 1}^{n_m} {\frac{{\partial {\mathcal{L}}({x_i},{y_i})}}{{\partial {z_i}(j)}}}  \cdot \frac{{\partial {z_i}(j)}}{{\partial {{\mathcal{W}}_{{{m,j}}}}}} \\
 &=\frac{1}{{bs}}\sum\limits_{i = 1}^{bs} {({y'_i}(j) - {y_i}(j))}  \sum\limits_{m = 1}^{n_m} \frac{{\partial {z_i}(j)}}{{\partial {{\mathcal{W}}_{{{m,j}}}}}},
\label{eq:eq4}
\end{split}
\end{equation}
where $\frac{{\partial {z_i}(j)}}{{\partial {{\mathcal{W}}_{{\rm{m,j}}}}}}=o_{m,i}$ is $m^{th}$ input of the fully-connected layer with an image $x_i$ and $n_m$ is the dimension number of the last hidden layer.
However, the above equation has two unknown values $y'_i{(j)}$ and $o_{m,i}$ in addition to the desired one $y_i(j)$.
We estimate the unknown values by feeding random samples to the classification model multiple times and derive sample numbers in each category $j$:
\begin{equation}
\sum\limits_{i = 1}^{bs} {{y_i}(j)} \approx f(\sum\limits_{i = 1}^{bs} {{{\tilde y'}_i}(j)}  - \frac{{bs \cdot \sum\limits_{m = 1}^{n_m} {\nabla _{{{\mathcal{W}}_{{\rm{m,j}}}}}}{\mathcal{L}}(\textbf{x},\textbf{y})}}{{{{\tilde o}_{i}}}}),
\label{eq:eq6}
\end{equation}
where ${\tilde o}_{i}=\sum\nolimits_{m = 1}^{n_m} mean({\tilde o}_{m,i})$, ${\tilde y'}_i$ and ${\tilde o}_{m,i}$ are generated from random samples, and $f(\cdot)$ is a mode function from multiple epochs to improve attacks.
Attackers obtain the gradient by using the uploaded classifier parameters, which can then be utilized to launch attacks and cause privacy leakage.

\subsection{Membership Inference Attack}
\label{MIA}

The adversary $C_A$ possesses a dataset $\mathcal{D}_{shadow}$ that includes some data records from the same distribution that are not in the target dataset.
This security assumption is strong, yet it maximizes detecting the protection of defenses.
The attack relies on overfitting caused by the trained classification model, which can be acquired from the submitted parameters.
Considering that overfitting has been mitigated by model aggregation in FL, we improve the attack by rebuilding different shadow models of a victim and other users.

\begin{figure}[t]
    \centering
    \includegraphics[width=1.0\linewidth]{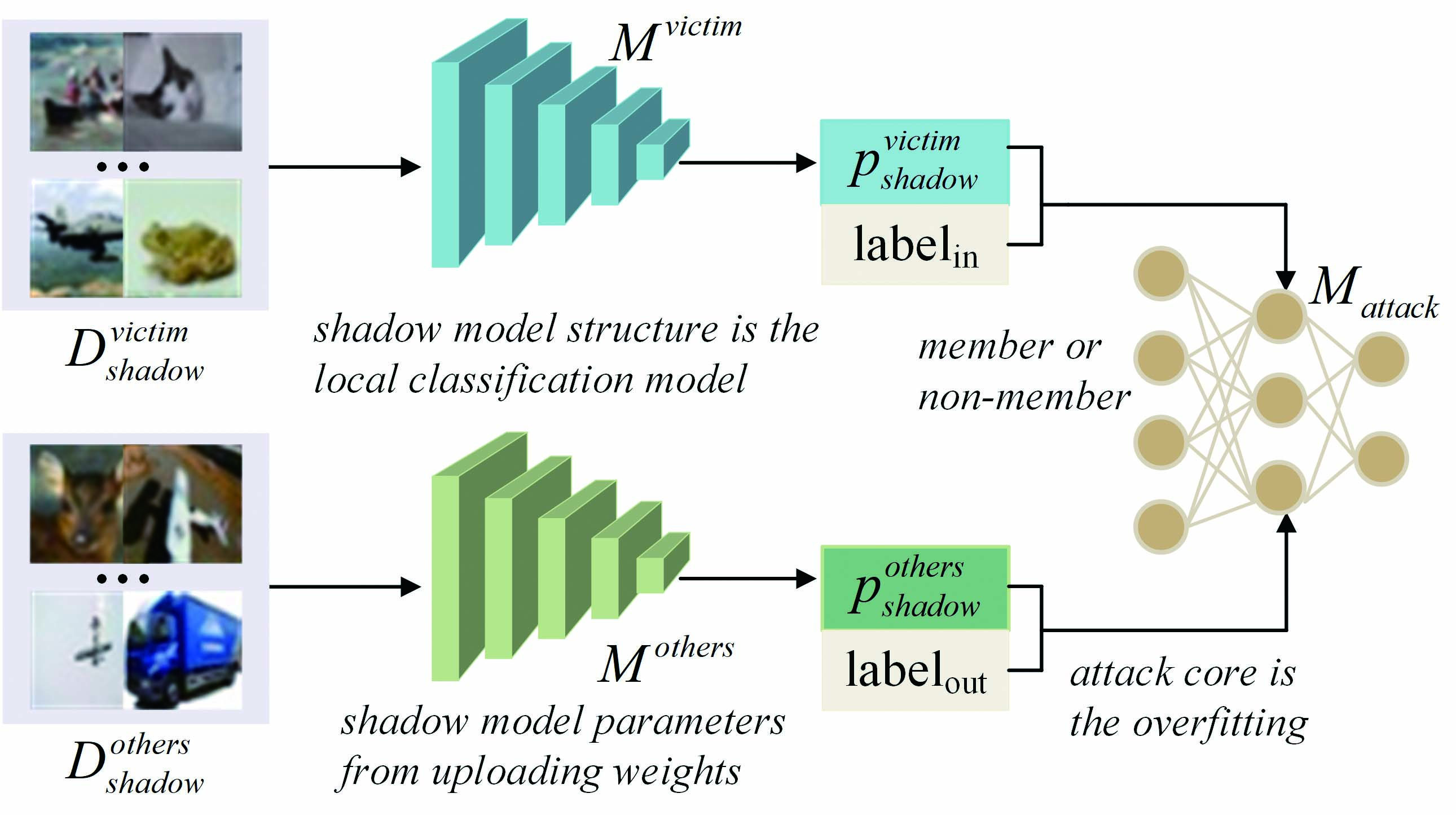}
    \caption{The enhanced membership inference attack.}
\label{fig:MIA_scheme}
\end{figure}

Given these uploaded model parameters of clients, $C_A$ produces copies of the victim model and other models, which we denote as $M^{victim}$ and $M^{others}$.
If client number $K>2$, $C_A$ aggregates all models of other users as $M^{others}$.
As depicted in Figure \ref{fig:MIA_scheme}, $C_A$ randomly divides the $\mathcal{D}_{shadow}$ into two disjoint sets: $\mathcal{D}_{shadow}^{victim}$ and $\mathcal{D}_{shadow}^{others}$, and then inputs them into $M^{victim}$ and $M^{others}$ respectively, generating the prediction vectors $p_{shadow}^{victim}$ and $p_{shadow}^{others}$.
The former is manually labeled $in$ (member) and the latter is labeled $out$ (non-member).
Next, $C_A$ trains the inference model $M_{attack}$ with ($p_{shadow}^{victim}$, $in$) and ($p_{shadow}^{others}$, $out$).
The attacker has a skeptical dataset and is unable to distinguish between data coming from the victim and other users.
Finally, the adversary feeds the skeptical dataset into trained $M_{attack}$, and the success rate is the percentage of correctly inferred data records. 

\subsection{Image Reconstruction Attack}
\label{RS}

Several image recovery optimization functions minimize the gradient difference between the victim uploaded gradient $\nabla \mathcal{W}$ and the generated gradient by dummy images $x^*$ with dummy or inferred labels $y^*$:
\begin{equation}
{x^*} = \arg \mathop {\min }\limits_{{x}} {\left\| {\frac{{\partial \mathcal{L}({x},{y^*};\mathcal{W})}}{{\partial \mathcal{W}}} - \nabla \mathcal{W}} \right\|^2}.
\label{eq:eq8}
\end{equation}

We conduct a theoretical investigation of these attacks described above, which are caused by sharing trained classifier parameters and result in privacy leakage.
In the Experiments section, we report detailed comparison attack results under the protection of our proposal and other defense methods.

\section{Methodology}
In PPIDSG, we aim to achieve three objectives:
1) Privacy Objective: safeguard users' privacy and resist attacks;
2) Parameter Objective: avoid trained classifier parameters from being uploaded when finishing global training in FL;
3) Utility Objective: maintain the model's classification accuracy.

\subsection{Framework of PPIDSG}

\begin{figure*}[t]
      \centering
      \includegraphics[width=1.0\linewidth]{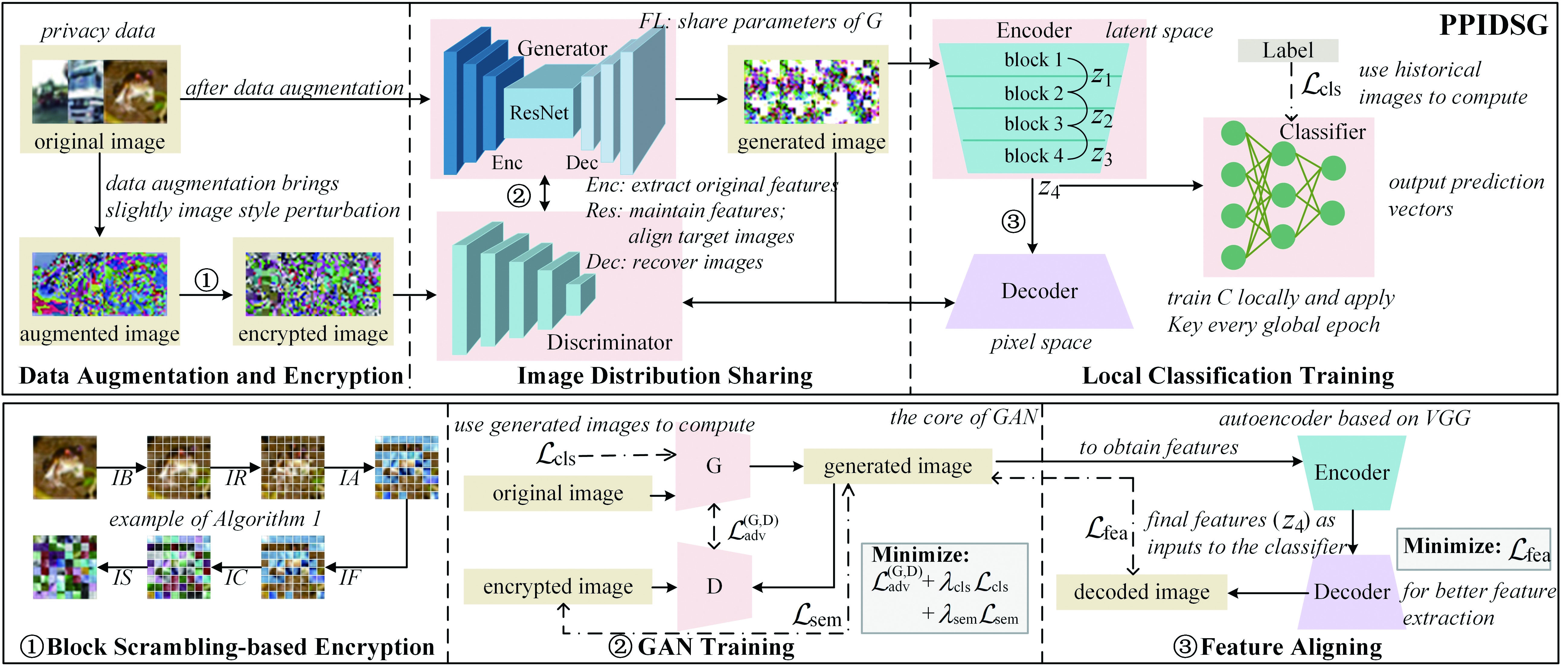}
\caption{The overview of our proposed framework.}
\label{fig:IEPS}
\end{figure*}

Figure \ref{fig:IEPS} illustrates our privacy-preserving framework which mainly consists of three modules: 1) a block scrambling-based encryption algorithm; 2) an image distribution sharing method; 3) local classification training.
To satisfy ``Privacy Objective'', we employ the encryption algorithm to convert original images into encrypted images (target domain) and train the classifier ($C$) locally.
Additionally, data augmentation \cite{augmentation} expands original samples while protecting privacy more effectively. 
To fulfill ``Parameter Objective'', we deploy the distribution sharing method based on CycleGAN by transmitting only the parameters of a generator ($G$) instead of $C$ to the server.
We train $G$ with a discriminator ($D$) to capture image distribution and eliminate the cycle consistency loss to prevent $G$ from reconstructing target images into original images.
To achieve ``Utility Objective'', we introduce a feature extractor ($F$) that utilizes an autoencoder \cite{Autoencoder} to extract interesting features and train it separately from $C$.
Then, a classification loss is applied to $G$ to focus on the distribution over image categories, for a better classification utility.

We consider a federated system with $K$ clients.
Each client $C_k$ has a local dataset $\mathcal D_k$. 
There are $N$ samples in total. 
Each client in our paper trains its models locally, uploading only the generator parameters $W$ to the central server.
Then, the server simply aggregates the model parameters in ${(t+1)}^{th}$ communication epoch by:

\begin{equation}
    {W^{t + 1}} \leftarrow \sum\limits_{k = 1}^K {\frac{{{n_k}}}{N}} ({W^t} - \alpha {\nabla {\cal L}({W^t};\mathcal{D}_{k})}),
\label{eq:FedAVG}
\end{equation}
where $\alpha$ and $\cal L$ represent the learning rate and loss function.

\subsection{Block Scrambling-based Encryption}
\label{encryption}

In this paper, we upload parameters of $G$ that may be utilized by an attacker to generate images that are similar to the target domain to execute attacks, thus we need to encrypt the target domain before training.
According to DeepEDN \cite{DeepEDN}, an image can be regarded as encrypted if it can be transformed into a domain that is quite dissimilar to the original.
While we can apply any transformation methods to encrypt images, we prefer the block scrambling-based encryption algorithm taking into account the model utility and encryption timeliness:

\begin{itemize}
    \item IR: The resulting block can be rotated 0, 90, 180, and 270 degrees. 
    \item IA: The L-bit pixel $p(i)$ can be adjusted with a pseudo-random bit $r_i$. The new pixel $\hat p(i)$ is calculated by
    \begin{equation}
    \begin{split}
     \hat p(i) = \left\{ {\begin{array}{*{20}{c}}
    {p(i),}&{{r_i} = 1}\\
    {p(i) \oplus ({2^L} - 1),}&{{r_i} = 0}
    \end{array}} \right..
    \end{split}
    \end{equation}
    \item IF: The resulting block can be flipped horizontally or vertically or not.
    \item IC: The colored image block can exchange pixel values in three color channels.
\end{itemize}
 
As illustrated in Algorithm \ref{alg:algorithm1}, the above operations are controlled by random keys.
We denote the distribution of the original image domain $X$ as $x_{i} \sim {p_{data}}(x)$ and the distribution of the encrypted image domain $\hat{X}$ as $\hat{x}_{i}\sim {p_{data}}(\hat{x})$.
Encrypted images are regarded as a target domain. 

\begin{algorithm}[ht]
\caption{Block Scrambling-based Encryption.}
\textbf{Input}: original image $x_{i}$, randomly key $K_j$, $j \in \{1, \cdots, 5\}$.\\
\textbf{Output}: encrypted image $\hat{x}_{i}$.
\begin{algorithmic}[1] 
\STATE IB($x_{i}$): $x_{i}$ with $P_x \times P_y$ pixels be divided into $n$ non-overlapped blocks ${\mathcal{B}}_l^{(0)}$ with $B_x \times B_y$ pixels, $l$ $\in$ $[1, n]$.
\FOR{each ${\mathcal{B}}_l^{(0)}$, $l$ $\in$ $[1, n]$}
\STATE ${\mathcal{B}}_l^{(1)}$ $\leftarrow$ IR(${\mathcal{B}}_l^{(0)}$, $K_1$);
\STATE ${\mathcal{B}}_l^{(2)}$ $\leftarrow$ IA(${\mathcal{B}}_l^{(1)}$, $K_2$);
\STATE ${\mathcal{B}}_l^{(3)}$ $\leftarrow$ IF(${\mathcal{B}}_l^{(2)}$, $K_3$);
\STATE ${\mathcal{B}}_l^{(4)}$ $\leftarrow$ IC(${\mathcal{B}}_l^{(3)}$, $K_4$); \verb|//|optional
\ENDFOR
\STATE IS(${\mathcal{B}}_l^{(4)}$, $K_5$): Shuffle and assemble all blocks ${\mathcal{B}}_l^{(4)}$ to generate a new encrypted image $\hat{x}_{i}$ with $K_5$.
\STATE \textbf{return} $\hat{x}_{i}$
\end{algorithmic}
\label{alg:algorithm1}
\end{algorithm}

\subsection{Image Distribution Sharing}
The fundamental idea of CycleGAN is to transfer original images into target images while capturing the target distribution.
Inspired by the idea, we utilize the structure to perform image distribution capture and upload parameters to finish FL.
Also, using the original image instead of noise as the original domain can further enhance privacy protection.
$G$ consists of three components: an encoder, ResNet blocks, and a decoder.
The encoder, consisting of convolutional layers, is employed to extract features from original images.
ResNet blocks are mainly responsible for maintaining original features and aligning them toward target images.
The decoder consists of deconvolution and convolutional layers, which restores the feature vector to the image.
By distinguishing generated images from target images, $D$ aims to improve the image distribution capture performance.
We apply an adversarial loss without conditional labels to train $G$ and $D$.
Given training samples $\{{x_{i}}\}_{i = 1}^{bs}$ and $\{{\hat{x}_{i}}\}_{i = 1}^{bs}$, the objective can be written as:
\begin{equation}
\begin{split}
{{\cal L}_{adv}^{\left( {G,D} \right)}} & = {{E}_{\hat{x}_i \sim {p_{data}}(\hat{x})}}\left[ {\log {D}(\{{\hat{x}_{i}}\}_{i = 1}^{bs})} \right] \\ 
&+ {{E}_{x_i \sim {p_{data}}(x)}}\left[ {\log(1-D(G(\{{x_{i}}\}_{i = 1}^{bs})))} \right].
\end{split}
\label{eq:GANloss}
\end{equation}

\begin{table*}[t]
\centering
\begin{tabular}{c|c|ccc|ccc|ccc}
\hline
{\textbf{Policy}} & {\textbf{Dataset}} & \multicolumn{3}{c|}{\textbf{ResNet18}} & \multicolumn{3}{c|}{\textbf{LeNet}} & \multicolumn{3}{c}{\textbf{ConvNet}} \\ \cline{3-11} 
                                 &                                   & Sigmoid   & ReLU    & LReLU  & Sigmoid   & ReLU    & LReLU  & Sigmoid   & ReLU    & LReLU  \\ \hline
{\textbf{ATS}}                   & MNIST                             & 100         & 98.44     & 96.88        & 100        & 98.44    & 100         & 98.44      & 79.69    & 70.31        \\
                                 & F-MNIST                           & 100      & 95.31      & 95.31          & 100        & 96.88    & 98.44       & 98.44      & 73.44    & 71.88        \\
                                 & CIFAR10                           & 100         & 93.75     & 98.44        & 100        & 92.19    & 92.19       & 95.31      & 62.50    & 60.94        \\
                                 & SVHN                              & 100         & 98.44     & 100          & 100        & 93.75    & 93.75       & 95.31      & 70.31    & 71.88        \\ \hline
{\textbf{EtC}}                   & MNIST                             & 100         & 100       & 100          & 100        & 95.31    & 93.75       & 93.75      & 98.44    & 96.88        \\
                                 & F-MNIST                           & 100      & 96.88     & 96.88          & 100        & 96.88    & 96.88       & 100      & 95.31    & 92.19        \\
                                 & CIFAR10                           & 100         & 96.88     & 96.88        & 100        & 96.88    & 96.88       & 100        & 92.19    & 85.94        \\
                                 & SVHN                              & 100         & 93.75     & 95.31        & 100        & 96.88    & 96.88       & 100        & 85.94    & 87.50        \\ \hline
\end{tabular}
\caption{The LIA accuracy (\%) of different model architectures and activation functions under the protection of ATS and EtC policy. We speculate that the reason for the low success rate in ConvNet is the MaxPool structure. LReLU: LeakyReLU.}
\label{tab:LIA1}
\end{table*}

G tries to minimize the objective and D tries to maximize it.
We add a $\ell_1$ norm loss function to further constrain semantic information.
The semantic loss is:
\begin{equation}
  {{\cal L}_{sem}} = \sum\nolimits_{i = 1}^{bs} \left\| {{{\rm{G}}_{X \to \hat{X}}}({x_i};{\theta _G}) - {\hat{x}_i}} \right\|. 
\label{eq:semloss}
\end{equation}

To lessen the negative effect caused by parameter collapse, we follow CycleGAN to update $D$ with a series of historical images rather than using new images from a separate epoch.
In addition, we add a classification loss computed by $C$ to $G$ to better focus on categorical information in the distribution.
Images generated by $G$ are gradually aligned toward the target distribution during training.
Then, we upload generator parameters to the central server to enable clients to complete FL.
This also facilitates the training of global $G$ because of more samples.

\subsection{Local Classification Training}

The feature extractor is introduced to achieve decoupling training in the non-overlapping and different images among clients.
$F$ is made up of an encoder $Enc$ and a decoder $Dec$.
We divide $Enc$ into four blocks $Enc_j$, where $j \in \{1,2,3,4\}$.
Each encoder block attempts to extract features, which are then restored to original dimensional pictures by the decoder. 
Let ${\tilde x}_i$ represents a generated image ${{\rm{G}}_{X \to \hat{X}}}({x_i};{\theta _G})$.
The objective of $F$ is to extract efficient features from generated images by minimizing
\begin{equation}
  {{\mathcal{L}}_{fea}} = \sum\nolimits_{{i} = 1}^{{bs}} {{{\left\| {Dec(Enc({{\tilde x}_i})) - {{\tilde x}_i}} \right\|}^2}}.
\label{eq:fealoss}
\end{equation}

Similarly, we use historical images to train $F$.
Then, we initialize $C$ and use the parameter as a key to guide the training of $G$ with the ${{\cal L}_{{{cls}}}}$ computed by $C$.
$C$ is initialized with the key before each beginning of the global communication round.
$C$ receives final features from $F$ and performs local update with $n_c$ classes by minimizing
\begin{equation}
  {{\cal L}_{{{cls}}}} =  - \sum\limits_{i = 1}^{bs} \sum\limits_{j = 1}^{n_c} {{y_{i}}} \left( j \right)\log { y'_{i}}\left( j \right). 
\label{eq:clsloss}
\end{equation}

Because $F$ converts encrypted images into efficient features, the classification network can be a fully connected neural network or a simple convolutional neural network.

\subsection{Full Objective}
\label{loss}
Above all, our full objective of GAN is
\begin{equation}
  {\cal L}_{GAN} = {{{\cal L}_{adv}^{\left( {G,D} \right)}} + {\lambda _{sem}}{{\cal L}_{sem}}} + {{\lambda _{cls}}{{\cal L}_{cls}}},
\label{eq:fullloss}
\end{equation}
where $\lambda$ controls the importance of each loss term.

\section{Experiments}
\label{experiment}

\subsection{Experiment Implementation}
\label{Experimental baselines}

\subsubsection{Datasets and Setup} 
We carry out experiments on four datasets: MNIST \cite{MNIST}, FMNIST \cite{f-mnist}, CIFAR10 \cite{CIFAR10}, and SVHN \cite{SVHN}.
Then we select one of the clients to act as a victim and carry out attacks.
Our experiments are performed on the PyTorch platform using an NVIDIA GeForce 3090 Ti GPU.

\subsubsection{Defense Baselines} 
We compare our method with several defenses in federated learning:
1) ATS \cite{privacy-preserving} suggests an automatic transformation search to find the ideal image transformation strategy;
2) EtC \cite{EtC} utilizes a block-based image transformation method to encrypt images;
3) DP \cite{DP} clips gradients and adds Gaussian noise during training;
4) GC \cite{DLG} performs privacy protection by pruning gradients;
5) FedCG \cite{FedCG} leverages conditional generative adversarial networks to guarantee privacy in federated learning.
With the privacy budget ${\epsilon}/T$ in $T$ global training epochs, we denote DP as DP\textless{}$\epsilon$, C\textgreater{}, where $C$ refers to the clipping bound.
We also explore GC with different gradient compression degrees.

\subsection{Attack Results}

\subsubsection{Results of Label Inference Attack}
We perform the attack (LIA) for 10 epochs under the protection of baselines (batch size is 64).
In these baselines, the activation function is commonly ReLU or Sigmoid.
We obtain the following observations from Table \ref{tab:LIA1} and Table \ref{tab:LIA2}:
1) attack success rates are essentially 100\% regardless of the complicated or simple models with Sigmoid;
2) rates are slightly lower, but most of them are more than 80\% in other activation functions.
One reason why Sigmoid performs better is that its results are all positive.
When users upload classifier parameters, especially the last few fully connected layers, an attacker can effectively perform label inference, despite any protection measures they have taken.
Compared with them, the attacker can't access penultimate gradients without sharing trained classifier parameters, making PPIDSG resist the attack.

\begin{table}[t]
\centering
\begin{tabular}{c|cc||cc||c}
\hline
                 & \textbf{DP1} & \textbf{DP2} & \textbf{GC1} & \textbf{GC2} & \textbf{FedCG}\\ \hline
{Sigmoid}     & 100          & 100          & 100          & 100          & 100     \\
{ReLU}      & 90.63        & 92.19        & 89.06        & 90.63        & 82.81   \\
{LReLU}     & 92.19        & 93.75        & 93.75        & 93.75        & 84.37   \\ \hline
\end{tabular}
\caption{The LIA accuracy (\%) under other protection policies in LeNet with the CIFAR10 dataset. DP1: DP\textless{}5,10\textgreater{}, DP2: DP\textless{}20,5\textgreater{}, GC1: GC (10\%), GC2: GC (40\%). More results are presented in the Appendix.}
\label{tab:LIA2}
\end{table}

\subsubsection{Results of Membership Inference Attack}

\begin{figure}[t]
      \centering
      \includegraphics[width=1.0\linewidth]{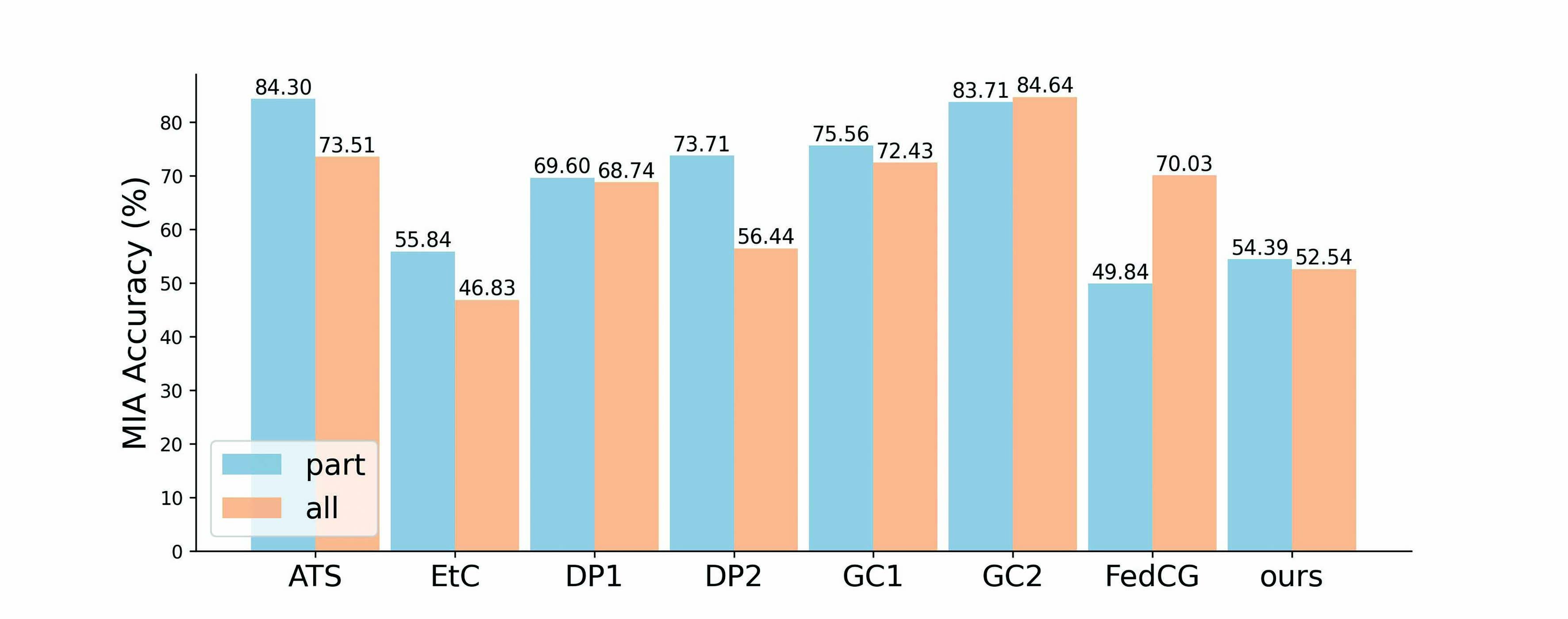}
\caption{The MIA accuracy (\%) of different victim data proportion with defense policies in the CIFAR10 dataset. Full dataset results are presented in the Appendix.}
\label{fig:MIA}
\end{figure}

We leverage test datasets as shadow datasets.
A fully connected network with layer sizes of 10, 128, and 2 (the output layer) is applied as $M_{attack}$.
We select the Adam optimizer and train the model for 100 epochs with a learning rate of 0.005.
We start by comparing with ML-Leaks \cite{ML-Leaks}, which exploits a single model for an attack, and uses two different data types as inputs to $M_{attack}$ in ATS.
As illustrated in Table \ref{tab:MIA1}, it is difficult to attack with a single model.
The observation demonstrates that our enhanced attack can improve the attack effect when overfitting is weakened in FL.
Additionally, we discover that the uniform image distribution prevents the image knowledge from attacking effectively.
Then we extend the attack to all defense strategies and assume that the attacker has a suspect dataset containing images from the victim and another user (part) or other users (all).
Figure \ref{fig:MIA} presents that only EtC and our method obtain lower attack effects than other defenses.
Uploading a trained classifier causes the exposure of overfitting, which further leads to privacy leakage.
We speculate that EtC achieves MIA resistance by encrypting the original image distribution.
Our proposal combines the above technique with the local classification model training to achieve privacy-preserving.

\begin{table}[t]
\centering
\begin{tabular}{c|c}
    \hline
    \textbf{Knowledge}                  & \textbf{MIA Accuracy (\%)} \\ \hline
    Prediction vectors (ML-Leaks)       & 51.92                     \\
    Prediction vectors (ours)           & 84.30                    \\ \hline
    original images (MIA)               & 45.81                    \\ \hline
\end{tabular}
\caption{Membership inference attack (MIA) accuracy using different data records and methods in the CIFAR10 dataset.}
\label{tab:MIA1}
\end{table}

\subsubsection{Results of Image Reconstruction Attack}
\label{RS_exp}

\begin{figure}[t]
      \centering
      \includegraphics[width=1.0\linewidth]{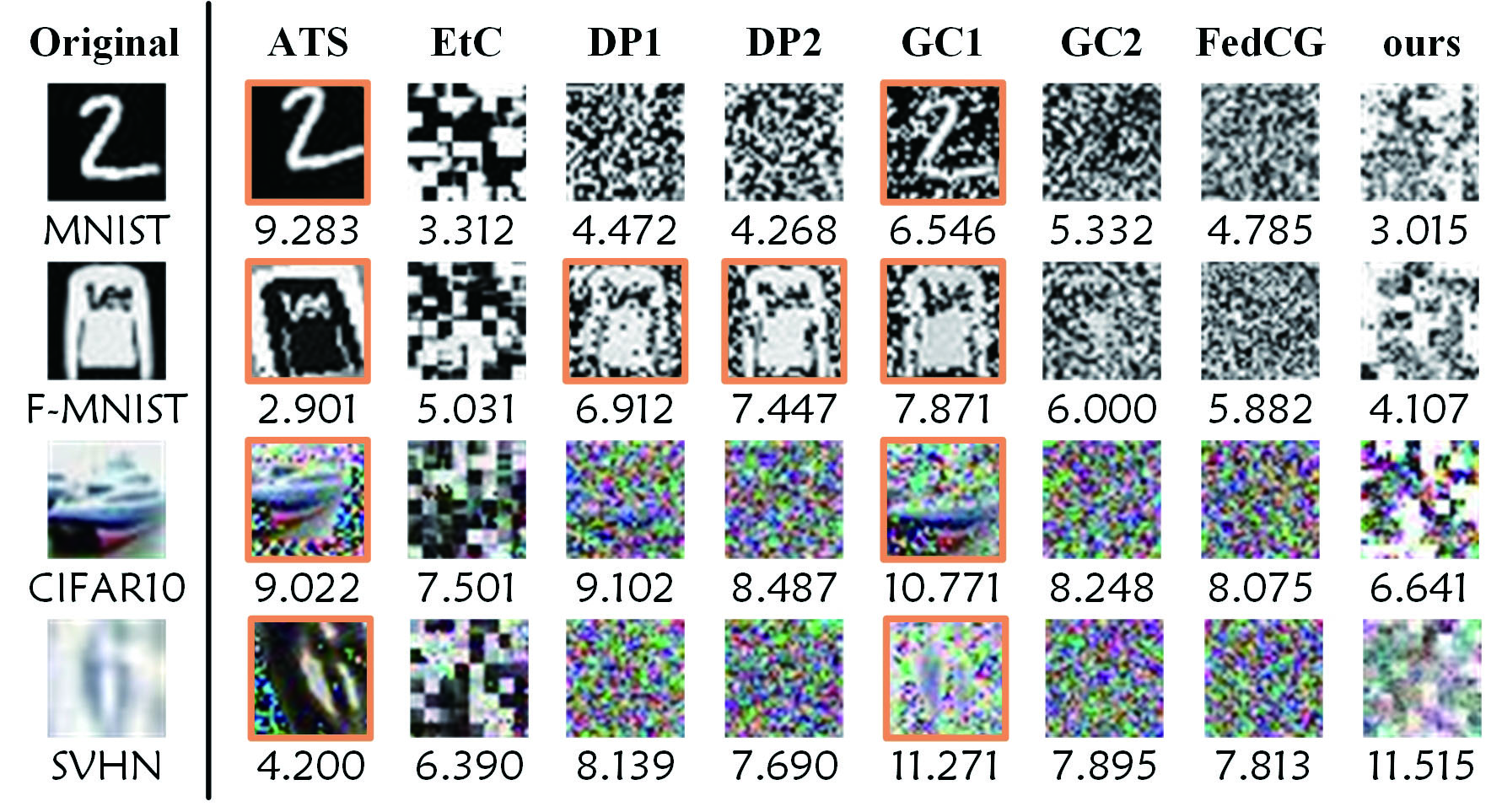}
\caption{Qualitative and quantitative results (PSNR: dB) of RS. Full results are shown in the Appendix.}
\label{fig:RS}
\end{figure}

Due to the lack of full model parameters and loss terms, the adversary fails to undertake this attack in PPIDSG and can only recover images from the outputs of $G$.
Figure \ref{fig:RS} visually shows evaluations of image reconstruction attacks (RS). 
Compared to ATS and GC (10\%), the attacker with other defense methods cannot recover original images visually. 
Moreover, we compare the privacy-preserving capability according to PSNR values between original and reconstructed images.
The lower the PSNR score, the higher the privacy-preserving of this defense policy.
We observe that PPIDSG achieves the lowest PSNR value in most datasets, implying the strongest privacy protection.
Although the value is high in the SVHN dataset, the attacker cannot visually restore images. 

\begin{table}[t]
\centering    
\begin{tabular}{c|cccc}
\hline
 \textbf{Policy}               & \multicolumn{4}{c}{\textbf{Classification Accuracy (\%)}}          \\ \cline{2-5} 
                                         & \textbf{MNIST} & \textbf{F-MNIST} & \textbf{CIFAR10} & \textbf{SVHN}  \\
    \hline
        ATS                              & 98.96          & 89.23            & 59.67            & 85.22          \\
        EtC                              & 98.06          & 89.41            & 53.34            & 78.70          \\
        DP1                              & 97.42          & 85.58            & 49.29            & 82.70          \\
        DP2                              & 97.54          & 85.01            & 44.43            & 80.28          \\
        GC1                              & 97.61          & 85.81            & 54.07            & 84.36          \\
        GC2                              & 97.22          & 85.09            & 50.91            & 79.96          \\
        FedCG                            & 98.60          & 88.00            & 53.20            & 79.71          \\
        ours                             & \textbf{99.43} & \textbf{91.60}   & \textbf{70.56}   & \textbf{91.53} \\
\hline
\end{tabular}
\caption{Classification accuracy results of test datasets. Full results are obtained in the Appendix.} 
\label{tab:acc}
\end{table}

By sharing classifier parameters to complete FL, not all baselines can protect against RS.
Additionally, these methods also perform poorly in LIA and MIA.
It indicates that sharing trained classifier parameters is a major privacy leakage in FL, and it may not be a suitable method for protection.

\subsection{Defense Performance}
\label{Defense_performance}

\subsubsection{Hyperparameter Configurations}
Our experiments are carried out on a FL system \cite{FedAVG} with ten clients, each of whom has the same amount of training data from the identical distribution.
We set batch size as 64, image pool size as 10, and block sizes $B_x$ and $B_y$ in the encryption algorithm are 4.
We apply an Adam optimizer and set the learning rate to 0.0002 in $G$ and $D$.
For $F$ and $C$, we use a SGD optimizer and set the learning rate to 0.01 (weight decay is 0.001). 
Their initial learning rates are constant in the first 20 global iterations and then decrease linearly until they converge to 0.
We set ${\lambda}_{sem}=1$, ${\lambda}_{cls}=2$ and run 50 global rounds.
A random user is selected for accuracy testing since there is no global classification model in PPIDSG.
More details can be found in the Appendix.

\begin{figure}[t]
    \centering
    \includegraphics[width=1\linewidth]{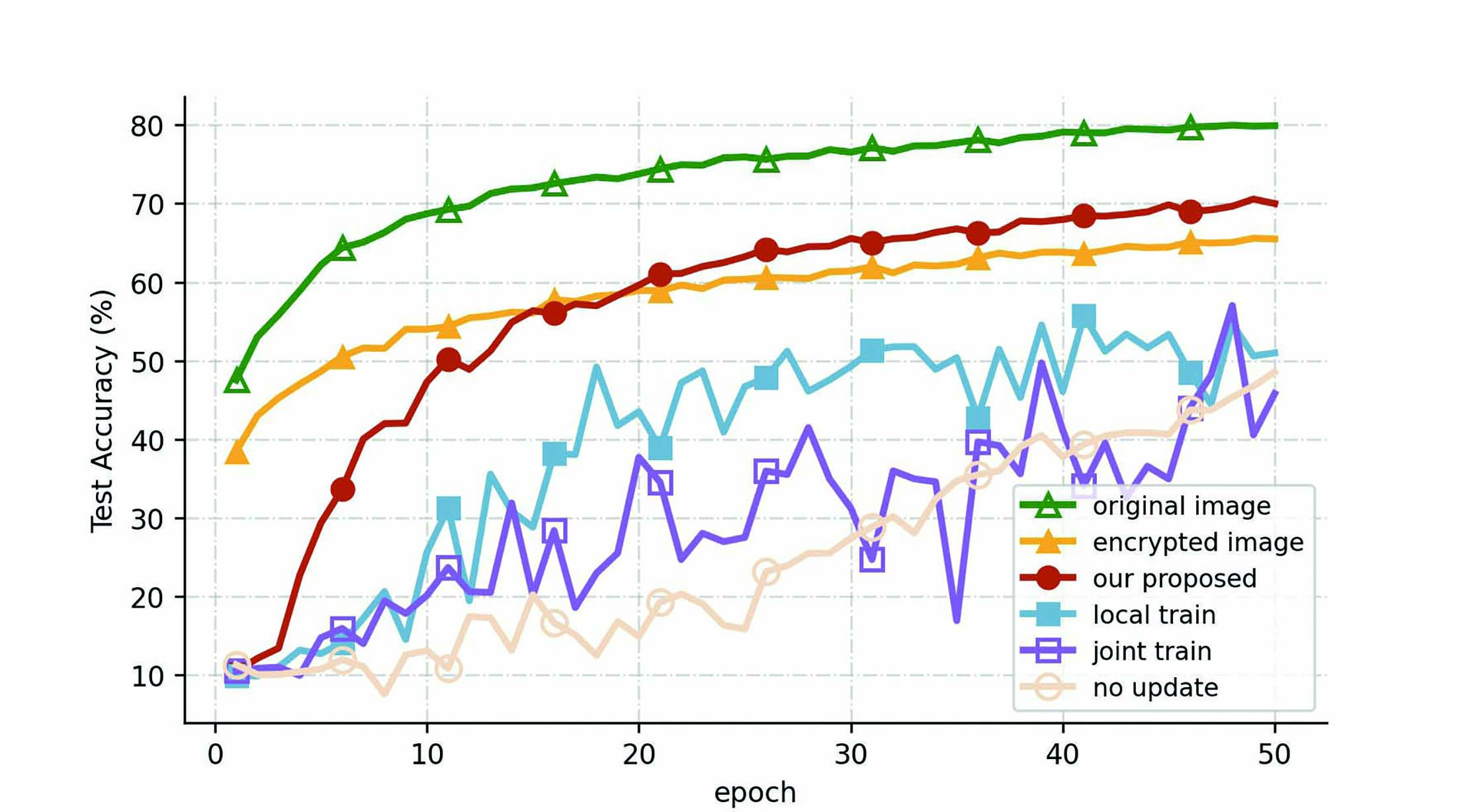}
\caption{Comparison of defense accuracy under different modules in the CIFAR10 dataset.}
\label{fig:acc_compare}
\end{figure}

\begin{figure}[t]
    \centering
    \includegraphics[width=1\linewidth]{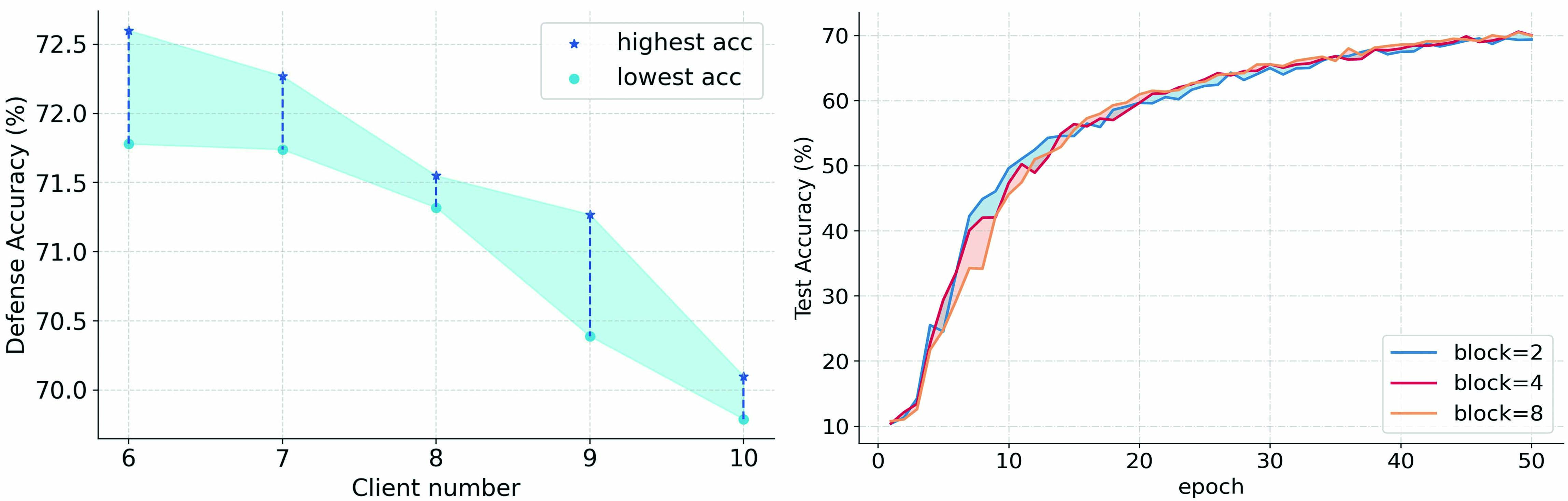}
\caption{Test accuracy of different client numbers with all clients (left) and different block sizes (right). More results are presented in the Appendix.}
\label{fig:ablation}
\end{figure}

\subsubsection{Results}
Table \ref{tab:acc} verifies the highest model classification accuracy of our method compared with other techniques, achieving the best utility in all datasets within a tolerable time overhead (see Appendix), especially in color images with an accuracy exceeding 70\%.
In addition, the classifier architecture complexity in ATS and EtC, which we exploit in this paper as ResNet18, greatly influences the model utility, whereas our feature extractor improves model performance while reducing classifier complexity.
In conclusion, our solution achieves competitive accuracy and outperforms others in terms of privacy-preserving.

To further investigate the model effectiveness of different modules, we conduct comparative studies as depicted in Figure \ref{fig:acc_compare}.
We have the following observations:
1) Compared with ``original image'' (benchmark) and ``encrypted image'' which take original images and encrypted images as inputs to $F$ which jointly train with $C$ respectively, our method has a slightly lower accuracy than the benchmark, exceeding the direct encryption policy.
This indicates our PPIDSG can extract useful features to maintain model utility, demonstrating that merely sharing the image distribution can finish federated learning.
2) Our approach is more stable than ``local train'' and ``joint train'', which train all models locally and jointly train $F$ with $C$, respectively.
It indicates our method can maintain a stable model while maintaining classification accuracy.
Also, it explains why we train $F$ and $C$ independently and share the parameter of $G$.
3) We also notice that ``no update'', which does not train the classifier and only uses the initial key, converges slowly.
Therefore, we only employ the initial key at the beginning of each global round.

We also explore the following aspects:
1) the impact of different client numbers;
2) the effect of various image block sizes.
As anticipated in Figure \ref{fig:ablation}, the block size has almost no effect on model convergence, while the user number has a slight effect on classification accuracy.

\begin{figure}[t]
	\centering
	\includegraphics[width=1\linewidth]{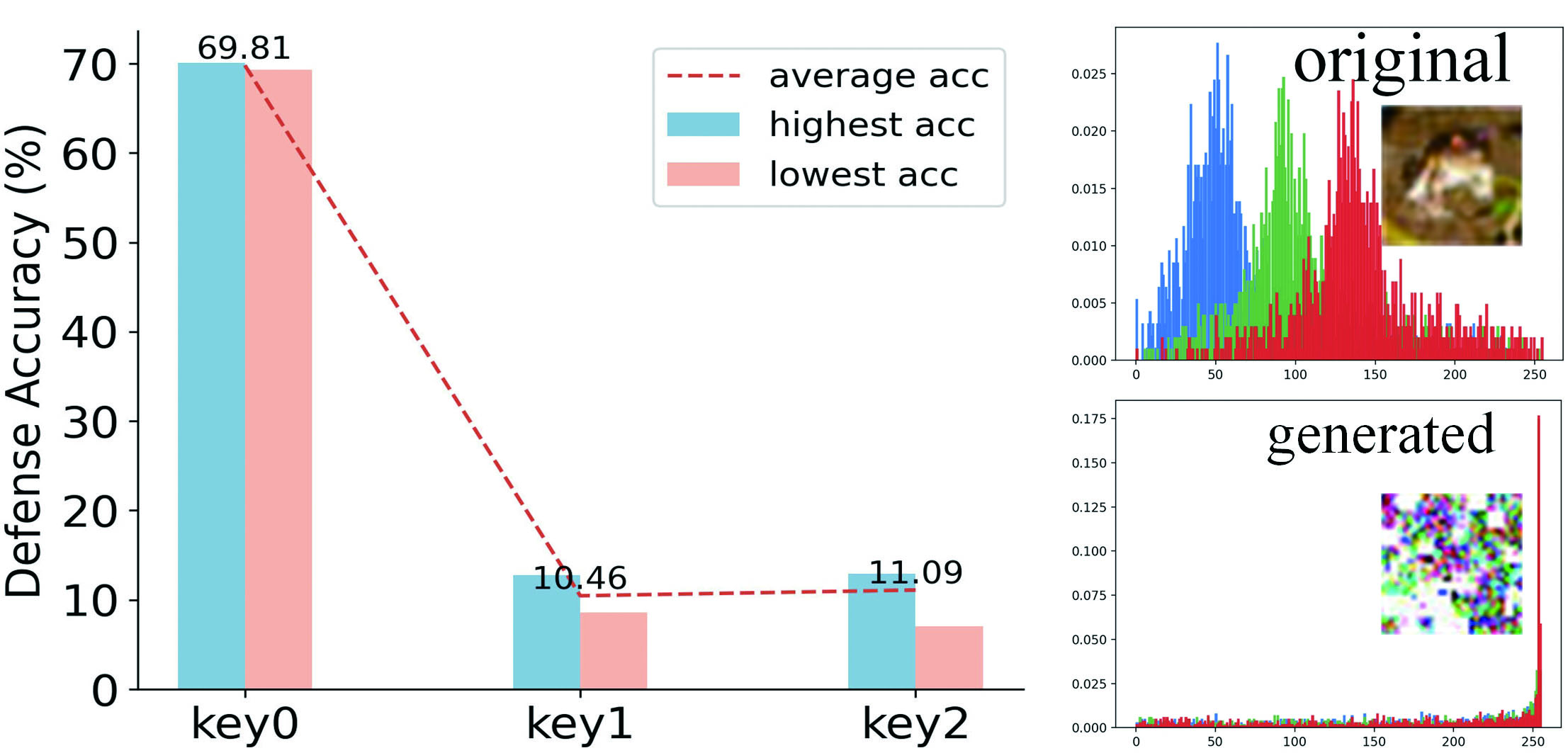}
\caption{Left: Test accuracy (\%) of classifier keys with all clients. Right: Pixel distribution between the original and generated images.}
\label{fig:key_disacc}
\end{figure}

\subsection{Security Analyses}
\label{security}

\subsubsection{Classifier Key}
The parameter of $C$, which has a total of 37764106 parameters in CIFAR10 and SVHN datasets and 29899786 parameters in MNIST and F-MNIST datasets, is a crucial privacy factor in the testing process.
We show how different parameters affect the test result by applying the corresponding $key0$ and two random keys.
In the left of Figure \ref{fig:key_disacc}, only $key0$ can perform valid testing while others cannot.

\subsubsection{Target Image}
The image space in the target domain determines the difficulty of using an exhaustive attack.
We formulate $N_{\rm{IR}}(n)$, $N_{\rm{IA}}(n)$, $N_{\rm{IF}}(n)$, and $N_{\rm{IC}}(n)$ as possible states of the encryption algorithm, and the target image space $N_{enc}{(n)}$ is represented by:
\begin{equation}
\begin{split}
  {{N}_{enc}}(n) &= {{N}_{\rm{IR}}}(n) \cdot {{N}_{\rm{IA}}}(n) \cdot {{N}_{\rm{IF}}}(n) \cdot {{N}_{\rm{IC}}}(n) \cdot n! \\
  & = {4^n} \cdot {2^n} \cdot {3^n} \cdot {6^n} \cdot n!,
\end{split}
\label{eq:key_space}
\end{equation}
where $n=({{P_x} \times {P_y}})/({{B_x} \times {B_y}})$, so that the target domain is a huge space that is difficult to decrypt.

\subsubsection{Generated Image}
We analyze the histograms of the original and generated images on the right of Figure \ref{fig:key_disacc}.
Sensitive data is protected visually, and the pixel distribution is more uniform statistically.
Particularly, blank areas of the generated image lead to abnormal distribution of final pixel areas, which we speculate can be compensated by improving the model aggregation and refining the image loss $\mathcal{L}_{sem}$.

\section{Conclusion}
We presented and supported the assumption that sharing trained classification model parameters is the main problem for privacy leakage in federated learning.
To solve the problem, we subsequently designed a novel privacy-preserving method (PPIDSG) that combines a block scrambling-based encryption algorithm, an image distribution sharing method, and local classification training.
Results showed that our scheme can successfully defend against attacks with high model utility.
Future work will focus on:
1) improving the model aggregation and stable training;
2) enhancing users' capacity to capture the image distribution.

\section{Acknowledgments}
This work was supported in part by the National Natural Science Foundation of China with Grant No. 62172383, No. 62231015, and No. 61802357, Anhui Provincial Key R\&D Program with Grant No. S202103a05020098, Research Launch Project of University of Science and Technology of China with Grant No. KY0110000049.
\bibliography{aaai24}

\clearpage
\appendix
\section{Appendix}
\section{Detail Supplement}

\subsection{Encryption Algorithm}
The target image domain in our paper consists of two components: 1) a block scrambling-based encryption algorithm; 2) the image style change due to the data augmentation.
For the encryption algorithm, we have six techniques:
\begin{itemize}
    \item{\emph{IB}}: We first divide the image (pixels: ${{{P_x} \times {P_y}}}$) into blocks (pixels: ${{{B_x} \times {B_y}}}$).
    In experiments, we default the image block size is 4 $\times$ 4.

    \item{\emph{IR}}: The resulting image block can be rotated 90, 180, and 270 degrees. There are four possible states, of which the fourth state is the image without rotating.

    \item{\emph{IA}}: The pixels in the resulting block either change or do not change. There are two possible states.
    
    \item{\emph{IF}}: The resulting image block can achieve horizontal and vertical flipping. The output block has three possible states, of which the third state is no flipping.
    
    \item{\emph{IC}}: The colored image block can exchange pixel values in three color channels, resulting in six possible states.
    
    \item{\emph{IS}}: All resulting blocks are shuffled and assembled, of which there are $n!$ ($n=\frac{{{P_x} \times {P_y}}}{{{B_x} \times {B_y}}}$) possibilities.
\end{itemize}

In color datasets such as the CIFAR10 dataset, we use the above six methods, while in black and white datasets like the MNIST dataset, we only use the first four methods and $\emph{IS}$.
In CIFAR10 and SVHN datasets, we use data augmentation (random rotation, horizontal flipping, and normalization) to restrain model collision and overfitting.
Since other datasets are relatively simple, we only use normalization.

\subsection{Label Inference Attack}
As shown in our paper, the prediction value $y'_i(j)=\frac{{{{\rm{e}}^{{{\rm{z}}_{\rm{i}}}{\rm{(j)}}}}}}{{\sum\nolimits_{j = 1}^{{n_c}} {{{\rm{e}}^{{{\rm{z}}_{\rm{i}}}{\rm{(j)}}}}} }}$, where $z_i$ is the final logits output before the softmax.
The cross-entropy loss function is widely adopted in most classification tasks (applicable to defense baselines and \emph{PPIDSG}), so the loss function can be written as:
\begin{equation}
{\mathcal{L}}(x_i,y_i) = - \sum\limits_{j = 1}^{{n_c}} {{y_i}(j)\log \frac{{{{\rm{e}}^{{{\rm{z}}_{\rm{i}}}{\rm{(j)}}}}}}{{\sum\nolimits_{j = 1}^{{n_c}} {{{\rm{e}}^{{{\rm{z}}_{\rm{i}}}{\rm{(j)}}}}} }}}.
\end{equation}
For each image $x_i$, the gradient w.r.t the network output without softmax at index $j$ is
\begin{equation}
\frac{{\mathcal L(x_i,y_i)}}{{\partial {z_i}(j)}} =  - \frac{{\partial {z_i}(c) - \log \sum\nolimits_{j = 1}^{{n_c}} {{e^{{z_i}(j)}}} }}{{\partial {z_i}(j)}} = {y'_i}(j) - {y_i}(j),
\end{equation}
if $j=c, y_i(j)=1$, else $y_i(j)=0$.
$c$ is the ground truth label.

\section{Network Architecture}
\subsection{Network of PPIDSG}

\subsubsection{GAN}
The generator $G$ is utilized for capturing the distribution of target images and we adopt 6 residual blocks for training.
Besides, we follow the naming convention used in \emph{CycleGAN}.
Let $cm$-$bni$-$k$ denotes a $m \times m$ Convolution-InstanceNorm-ReLU layer with stride $n$, padding $i$, and $k$ filters.
$pad$-$i$ denotes the ReflectionPad layer with padding $i$.
$dm$-$bni$-$k$ denotes a $m \times m$ ConvTranspose-InstanceNorm-ReLU layer with stride $n$, padding $i$, output padding $i$, and $k$ filters.
$t$-$k$ denotes a $7 \times 7$ Convolution-Tanh layer with stride $1$, padding $0$, and $k$ filters.
The residual block $Rk$ contains two reflection padding 1, two $3 \times 3$ convolution layers with stride 1 and padding 0, two instanceNorm blocks with $k$ filters, and one ReLU function. 

The structure of $G$ in CIFAR10 and SVHN datasets consists of: $pad$-$3$, $c7$-$b10$-$32$, $c3$-$b21$-$64$, $c3$-$b21$-$128$, $R128$, $R128$, $R128$, $R128$, $R128$, $R128$, $d3$-$b21$-$64$, $d3$-$b21$-$32$, $pad$-$3$, $t$-$3$.
Table \ref{param_num} shows the analysis of generator parameters in CIFAR10 and SVHN datasets, with a total of 1963200 parameters.
The only difference of MNIST and F-MNIST datasets is that the input and output channels are both one.
There has enough space to initialize the generator.

\begin{table*}[t]
\centering
\scalebox{0.85}{
\begin{tabular}{|c|c|c|c|c|c|c|}
\hline
\textbf{Layer}       & \textbf{Number} & \textbf{Size}  & \textbf{Input} & \textbf{Output} & \textbf{Parameters} & \textbf{Total Parameters}  \\ \hline
ConvBlock  & 1      & 7 $\times$ 7  & 3     & 32     & 4704       & 4704    \\ \hline
ConvBlock  & 1      & 3 $\times$ 3  & 32    & 64     & 18432      & 23136   \\ \hline
ConvBlock  & 1      & 3 $\times$ 3  & 64    & 128    & 73728      & 96864   \\ \hline
ResnetBlock & 6      & 3 $\times$ 3  & 128   & 128    & 1769472     & 1866336  \\ \hline
DeconvBlock & 1      & 3 $\times$ 3  & 128   & 64     & 73728      & 1940064 \\ \hline
DeconvBlock & 1      & 3 $\times$ 3  & 64    & 32     & 18432      & 1958496 \\ \hline
ConvBlock & 1      & 7 $\times$ 7  & 32    & 3      & 4704       & 1963200 \\ \hline
\end{tabular}
}
\caption{Parameter numbers of our generator.}
\label{param_num}
\end{table*}

\begin{figure}[t]
\centering
\centering\includegraphics[width=1.0\linewidth]{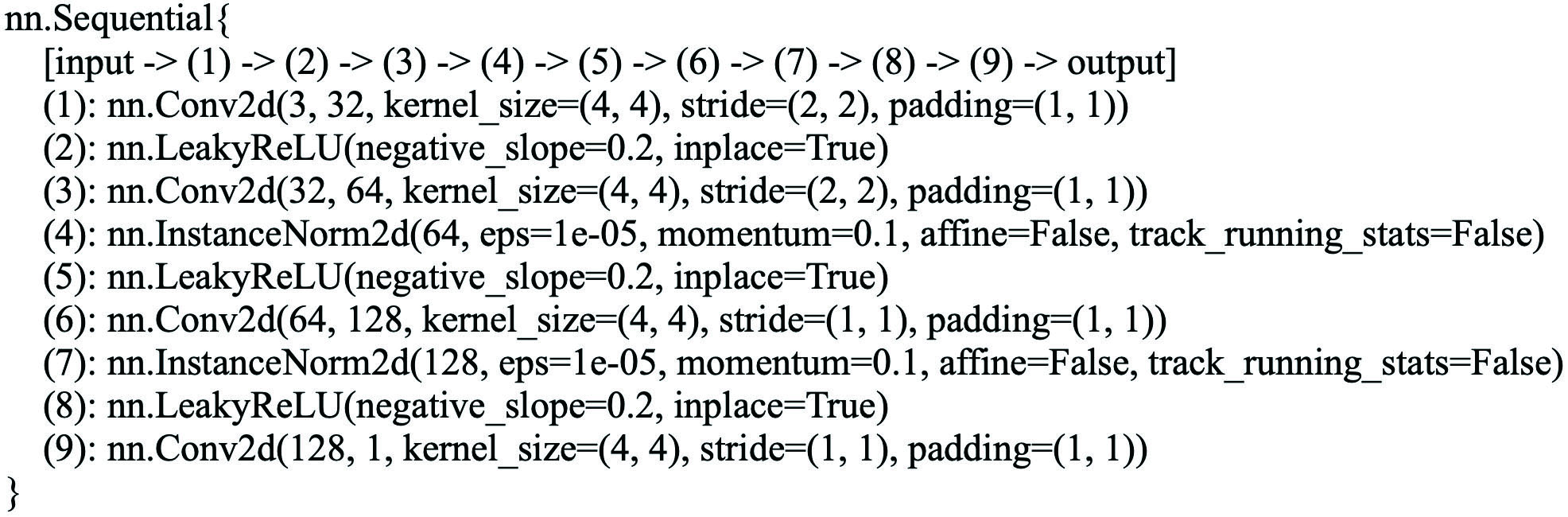}
\caption{The discriminator architecture of CIFAR10 and SVHN datasets.}
\label{Discriminator}
\end{figure}

The structure of discriminator $D$ in CIFAR10 and SVHN datasets is shown in Figure \ref{Discriminator}.
In MNIST and F-MNIST datasets, we just change the number of input channel to 1 and resize the image size to $28 \times 28$.

\subsubsection{Feature Extractor}
The structure of feature extractor $F$ in CIFAR10 and SVHN datasets is shown in Figure \ref{net1}.
Different from GAN, the feature extractor uses the BatchNorm layer.
So, $c$-$k$ denotes a $3 \times 3$ Convolution-BatchNorm-ReLU layer with stride $1$, padding $1$, and $k$ filters.
$M$ denotes a $2 \times 2$ MaxPool layer with stride $2$, padding $0$.
$d$-$k$ denotes a $4 \times 4$ ConvTranspose-BatchNorm-ReLU layer with stride $2$, padding $1$, and $k$ filters.
Figure \ref{net2} depicts the architecture of $F$ in MNIST and F-MNIST datasets.
The only difference is that the input and output channels are both one.

\subsubsection{Classifier}
\begin{figure}[t]
\centering
\centering\includegraphics[width=.8\linewidth]{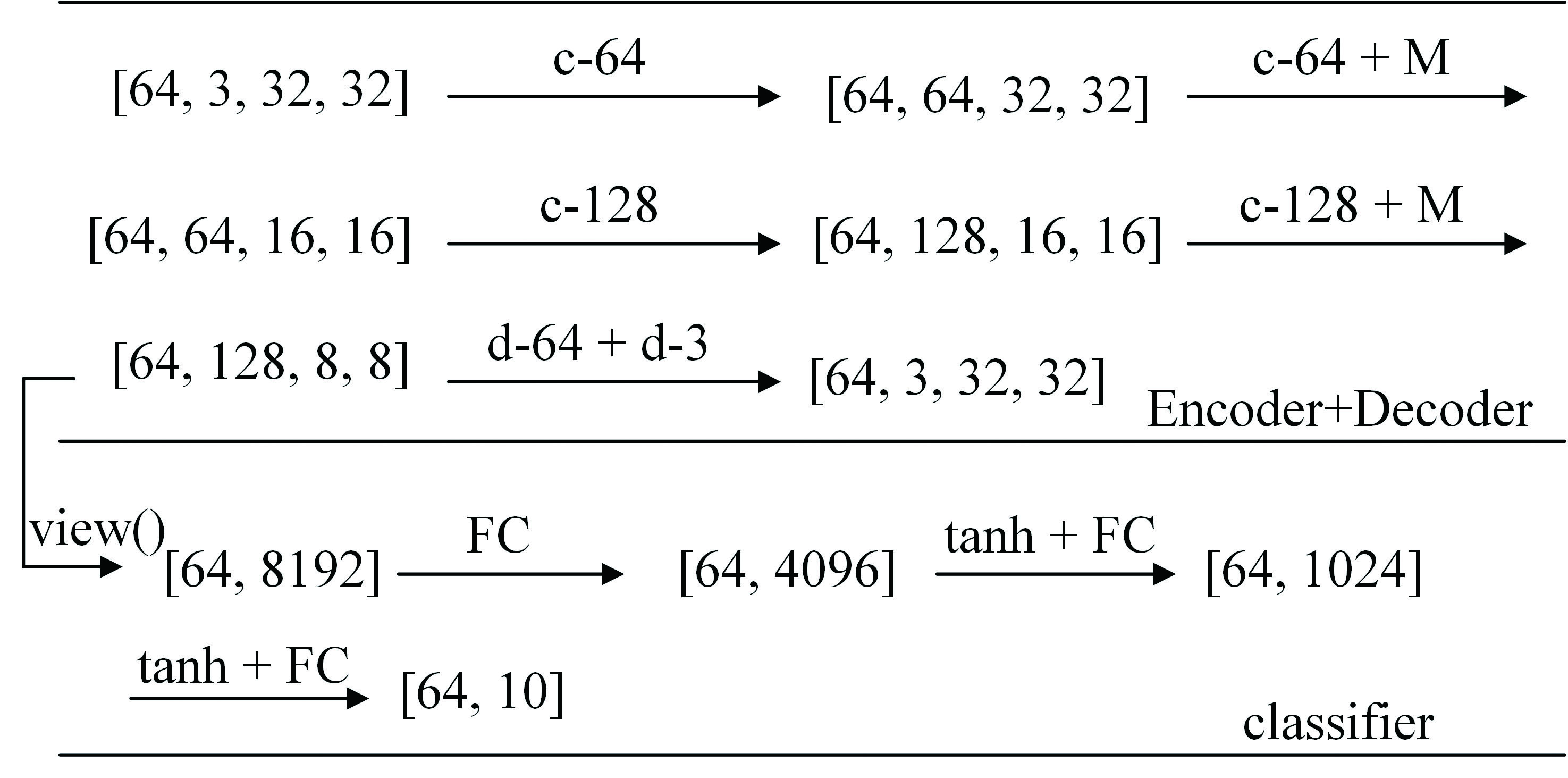}
\caption{Architecture of the feature extractor and classifier in CIFAR10 and SVHN datasets.}
\label{net1}
\end{figure}

\begin{figure}[t]
\centering
\centering\includegraphics[width=.8\linewidth]{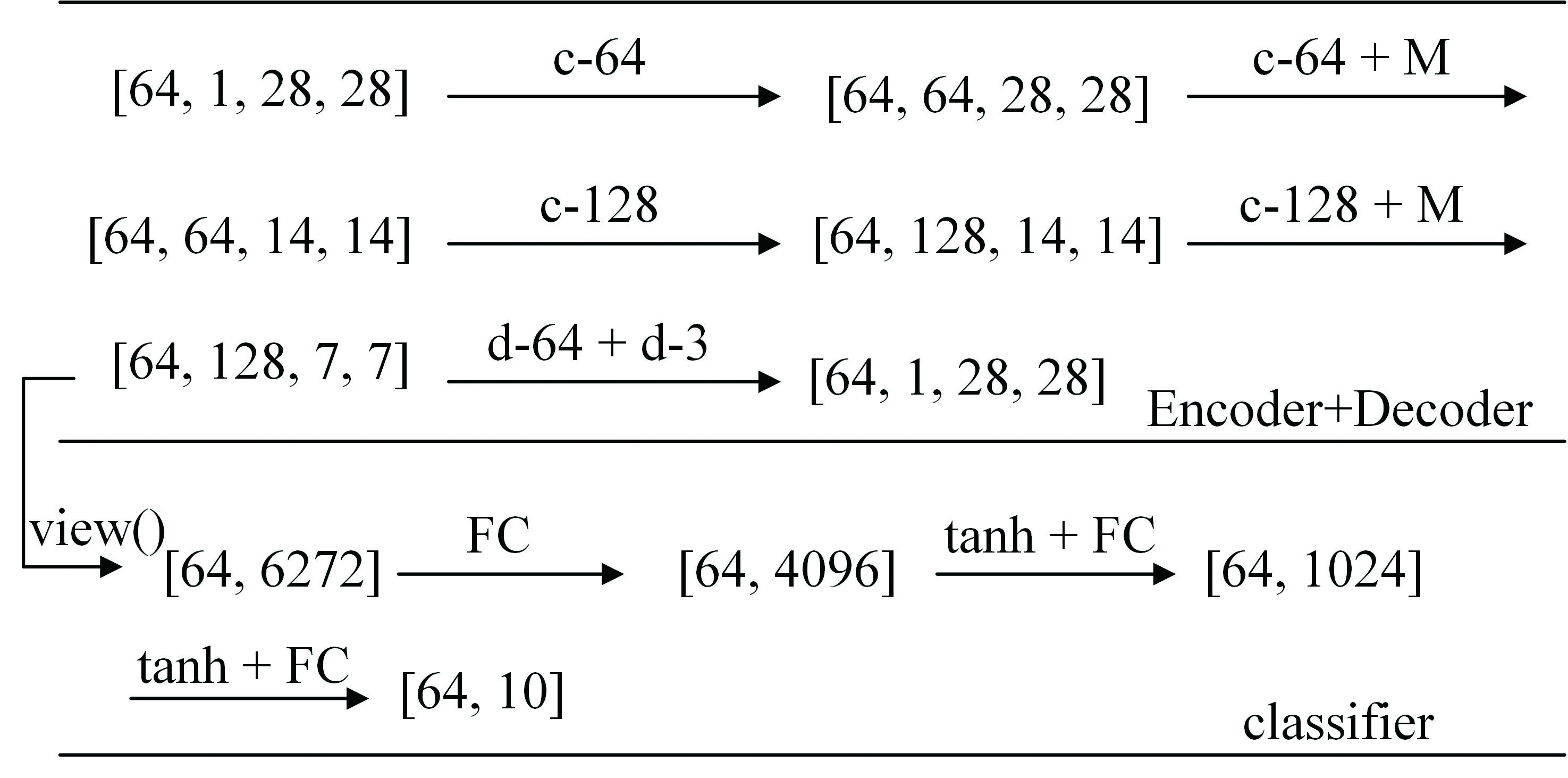}
\caption{Architecture of the feature extractor and classifier in MNIST and F-MNIST datasets.}
\label{net2}
\end{figure}

The classifier $C$ receives final features and performs local train.
In our paper, we use a simple fully connected neural network for classification.
The specific parameters are shown in Figure \ref{net1} and Figure \ref{net2}.

\subsection{Network of Baselines}

In the defense baselines, we mainly use two different classification networks: ResNet18 and LeNet.
Since the pixel space of an input image is 32 $\times$ 32 or 28 $\times$ 28, we made a simple modification to the classifier to make it suitable for datasets.
Based on ResNet18, we alter the kernel of the first convolutional network layer from 7 to 3 and remove the maximum pooling layer, keeping the rest unchanged.
For LeNet architecture, we refer to the structure in \emph{DLG} and modify the activation function from the Sigmoid function to the ReLU function to raise the classification accuracy in defense baselines.

In the policy of \emph{ATS} and \emph{EtC}, we choose various networks for experiments due to the diversity of model structures in the original text.
In the policy of \emph{DP} and \emph{GC}, LeNet is chosen as the classifier because the original and most of the related papers chose it as the classifier.
For \emph{FedCG}, the literature specifies the network structure of each component, and we follow the original setup.

In label inference attacks, we additionally use ConvNet to perform label inference in \emph{ATS} and \emph{EtC} policy.
The structure of ConvNet is $c$-$8$, $c$-$16$, $c$-$16$, $c$-$32$, two Maxpool layers, and a fully connected layer with ten out features.
More experimental details can be found in the source code.
In membership inference attacks, the model structures of $M^{victim}$ and $M^{others}$ are consistent with user classifiers.
The attack model $M_{attack}$ is a simple binary classifier whose input is the combination of labels and prediction vectors after softmax.
In image reconstruction attacks, we change the activation function from ReLU to Sigmoid because of the twice-differentiable model algorithm according to \emph{DLG}.

\subsection{The procedure of PPIDSG}

\begin{table*}[t]
\centering
\begin{tabular}{c|cccc||c|cccc}
\hline
{\textbf{Policy}} & \multicolumn{4}{c||}{\textbf{Accuracy(\%)}}                           & {\textbf{Policy}} & \multicolumn{4}{c}{\textbf{Accuracy(\%)}}                             \\ \cline{2-5} \cline{7-10} 
                                 & \textbf{MNIST} & \textbf{F-MNIST} & \textbf{CIFAR10} & \textbf{SVHN} &                                  & \textbf{MNIST} & \textbf{F-MNIST} & \textbf{CIFAR10} & \textbf{SVHN}  \\ \hline
\textbf{ATS}                     & 98.96          & 89.23            & 59.67            & 85.22         & \textbf{DP1}                     & 97.42          & 85.58            & 49.29            & 82.70          \\ \cline{1-5}
\textbf{EtC}                     & 98.06          & 89.41            & 53.34            & 78.70         & \textbf{DP2}                     & 97.40          & 85.71            & 49.61            & 79.25          \\ \cline{1-5}
\textbf{GC1}                     & 97.61          & 85.81            & 54.07            & 84.36         & \textbf{DP3}                     & 97.60          & 85.50            & 50.12            & 83.67          \\
\textbf{GC2}                     & 97.44          & 86.11            & 52.78            & 80.18         & \textbf{DP4}                     & 97.54          & 85.01            & 44.43            & 80.28          \\
\textbf{GC3}                     & 97.22          & 85.09            & 50.91            & 79.96         & \textbf{DP5}                     & 97.82          & 85.94            & 53.75            & 83.89          \\ \hline
\textbf{FedCG}                   & 98.60          & 88.00            & 53.20            & 79.71         & \textbf{ours}                    & \textbf{99.43} & \textbf{91.60}   & \textbf{70.56}   & \textbf{91.53} \\ \hline
\end{tabular}
\caption{Defense classification accuracy of victim model with various techniques.}
\label{tab:Defense_acc}
\end{table*}

In the local device, the user first performs data augmentation on original images $X$ and then encrypts them by using a block scrambling-based encryption algorithm to form the target domain $\hat{X}$.
Then, the user feeds these augmented images into $G$ to generate images ${\tilde X}$.
Encrypted images $\hat{X}$ and generated images ${\tilde X}$ are sent to $D$ for distribution capture, then obtain ${{{\cal L}_{adv}^{\left( {G,D} \right)}}}$ with G.
${\tilde X}$ are then sent to $F$ and $C$ for classification, and the cross-entropy loss ${{\cal L}_{cls}}$ is performed with labels.
At the same time, L1 loss ${{\cal L}_{sem}}$ is compared between ${\tilde X}$ and $\hat{X}$ to enhance the semantic information of the generated image.
$F$ and $C$ perform decoupling local classification training. 
$F$ compares ${\tilde X}$ with the image generated by the decoder in $F$, generates MSE loss ${{\mathcal{L}}_{fea}}$ to train.
After completing the local training, users upload the parameters of G for distribution sharing in a federated learning way.

For the central server, only the generator parameter is sent to the server to finish federated learning instead of a classifier for privacy. 
The server uses the parameter average to complete the model aggregation and sends the global parameter to users. 
Users employ it as their new generator to finish the next local training and then upload updated parameters until finishing train epochs.

\section{Experiments}

\begin{table}[t]
\centering
\scalebox{0.85}{
\begin{tabular}{c|c}
    \hline
{\textbf{Loss Term}}                                                        & \textbf{Accuracy(\%)} \\ \hline
\textbf{$\mathcal{L}_{GAN}$}                                                & 70.56           \\ \hline
\textbf{$\mathcal{L}_{GAN}-\mathcal{L}_{sem}$}                              & 69.19           \\ \hline
\textbf{$\mathcal{L}_{GAN}-\mathcal{L}_{cls}$}                              & 10.88           \\ \hline
\textbf{$\mathcal{L}_{GAN}-\mathcal{L}_{cls}-\mathcal{L}_{sem}$}            & 10.00           \\ \hline
\end{tabular}
}
\caption{Accuracy results of ablation study in the CIFAR10 dataset. Experimental results show that $\mathcal{L}_{cls}$ is a necessary term to complete the classification task. This is because we do not upload classifier parameters and thus can only use the initialized classifier for accuracy testing. And the generator cannot extract valid classification features without adding $\mathcal{L}_{cls}$ to G. $\mathcal{L}_{sem}$ can slightly enhance the model utility.}
\label{tab:acc_loss}
\end{table}

\begin{table}[t]
\centering
\scalebox{0.75}{
\begin{tabular}{c|c||c|c||c|c||c|c}
\hline
IR & 71.35 & IC     & 70.77 & IF+IC    & 70.55 & IR+IA+IC & 70.73 \\ \hline
IA & 70.57 & w/o IS & 70.06 & IR+IF    & 70.70 & IR+IF+IC & 70.09 \\ \hline
IF & 70.17 & IR+IA  & 69.59 & IR+IA+IF & 68.99 & IA+IF+IC & 71.00 \\ \hline
\end{tabular}
}
\caption{\small Model accuracy (\%). All contain IS except w/o IS.}
\label{tab:EAchoice}
\end{table}

\subsection{Defense Result}

We present the classification accuracy results of all defense measures in Table \ref{tab:Defense_acc}.
Because the classification accuracy of ATS and EtC policy greatly depends on the model structure, we use ResNet18 as their classification model.
In addition, the model utility of ATS is also related to its randomly selected transformation techniques.
Other baselines use LeNet as the classification model.
For the time analysis, PPIDSG maintains good utility within a tolerable time overhead.
For example, one epoch in PPIDSG is about 3 minutes in CIFAR10, which is faster than FedCG (30 minutes) and slower than other baselines (within 1 minute), but their time increases with the model complexity. 

\subsubsection{Ablation Study}
We explore the effects of different additional loss terms on the generator.
The results are shown in Table \ref{tab:acc_loss}.
We also extend the ablation study on encryption algorithm choice in Table \ref{tab:EAchoice} (CIFAR10 dataset). 
As the encryption algorithm becomes more complex, generated images contain less information and are more difficult to classify, so the performance decreases.

\begin{figure}[t]
    \centering
    \includegraphics[width=.8\linewidth]{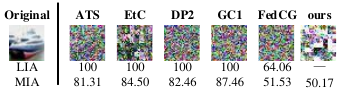}
    \caption{Attack results in trade-off model accuracy (40\%).}
\label{fig:tradeoff}
\end{figure}

\begin{figure}[t]
    \centering
    \includegraphics[width=.8\linewidth]{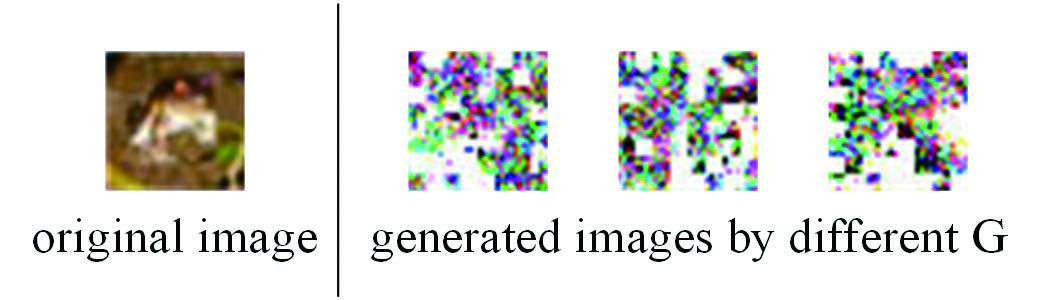}
\caption{The generated images by different generators.}
\label{fig:keyresult}
\end{figure}

Then, we explore the trade-off of our method compared to baselines.
The experiment result is shown in Figure \ref{fig:tradeoff}.

\subsubsection{Generated Image}
We initialize $G$ three times with the same setting and obtain different networks.
It is clear that the outputs obtained by feeding the same input from three generators are not similar in Figure \ref{fig:keyresult}.

\subsection{Label Inference Attack}
The attack accuracy is the percentage of successfully inferred labels out of all labels.
In case two modes appear at the same time, we pick the smaller one.
Table \ref{LIA_DP} and Table \ref{LIA_FedCG} show supplementary experimental results.

\subsection{Membership Inference Attack}

\begin{figure*}[t]
    \centering
    \includegraphics[width=1.0\linewidth]{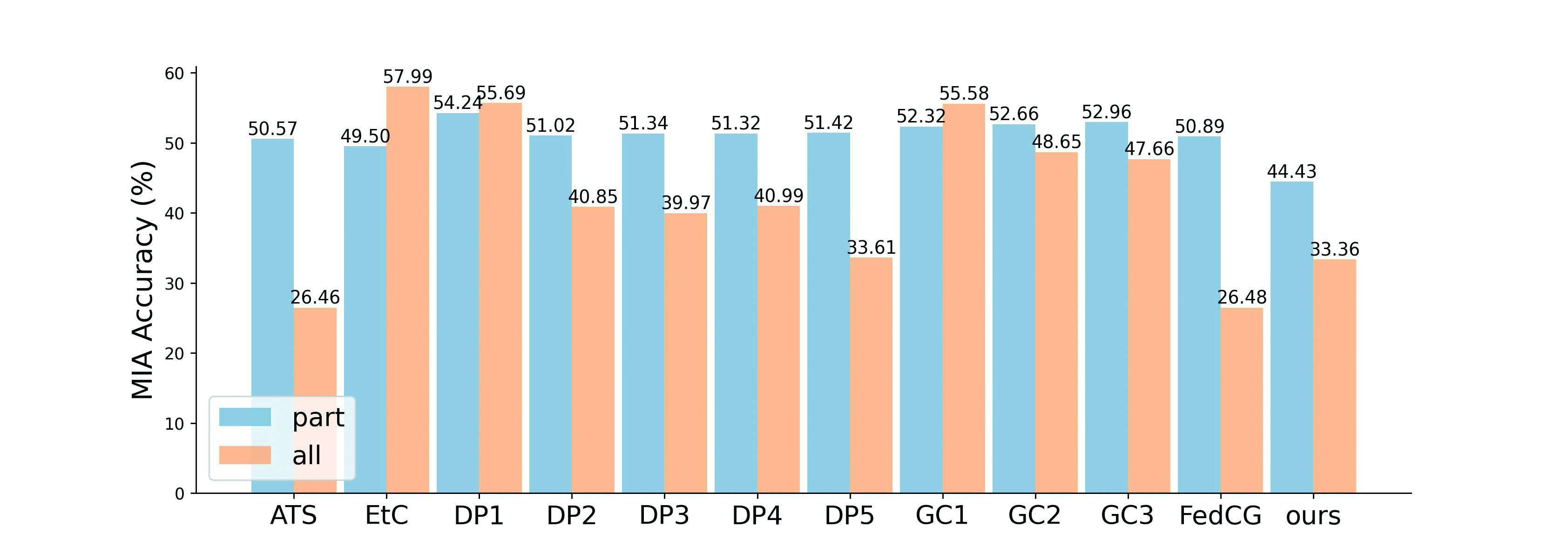}
    \caption{The MIA accuracy of different data partition under different defense policies in the MNIST dataset. DP1, DP2, DP3, DP4, DP5, GC1, GC2, and GC3 represent DP\textless{}5,10\textgreater{}, DP\textless{}10,10\textgreater{}, DP\textless{}20,10\textgreater{}, DP\textless{}20,5\textgreater{}, DP\textless{}20,20\textgreater{}, GC(10\%), GC(20\%), and GC(40\%). The success rates are relatively low. All this is owing to the dataset being composed of numbers with the same general characteristics, which are more difficult to attack. Our method is still able to achieve a low success rate among all defense methods.}
\label{fig:MIA_MNIST}
\end{figure*}

\begin{figure*}[t]
    \centering
    \includegraphics[width=1.0\linewidth]{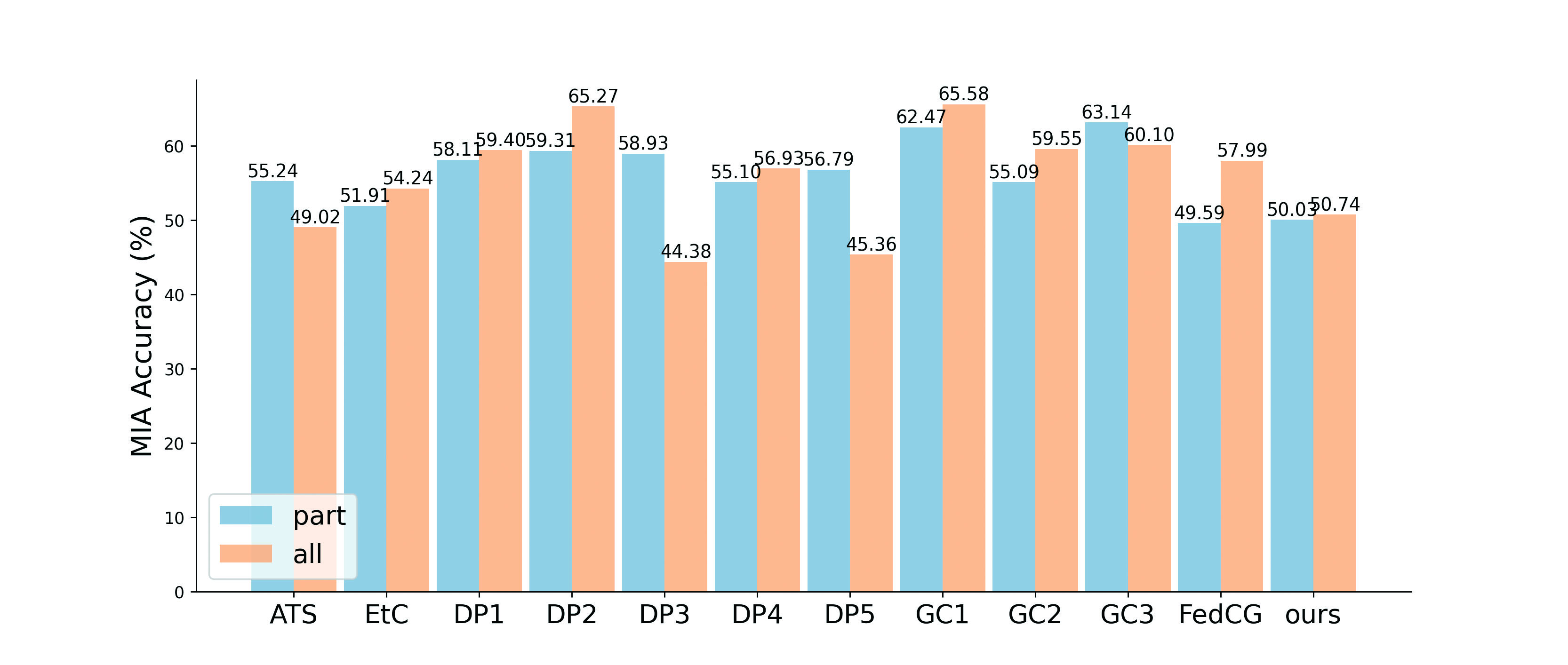}
    \caption{The MIA accuracy of different data partition under different defense policies in the Fashion-MNIST dataset. Our method is still able to achieve a low success rate among all defense methods.}
\label{fig:MIA_FMNIST}
\end{figure*}

\begin{figure*}[t]
    \centering
    \includegraphics[width=1.0\linewidth]{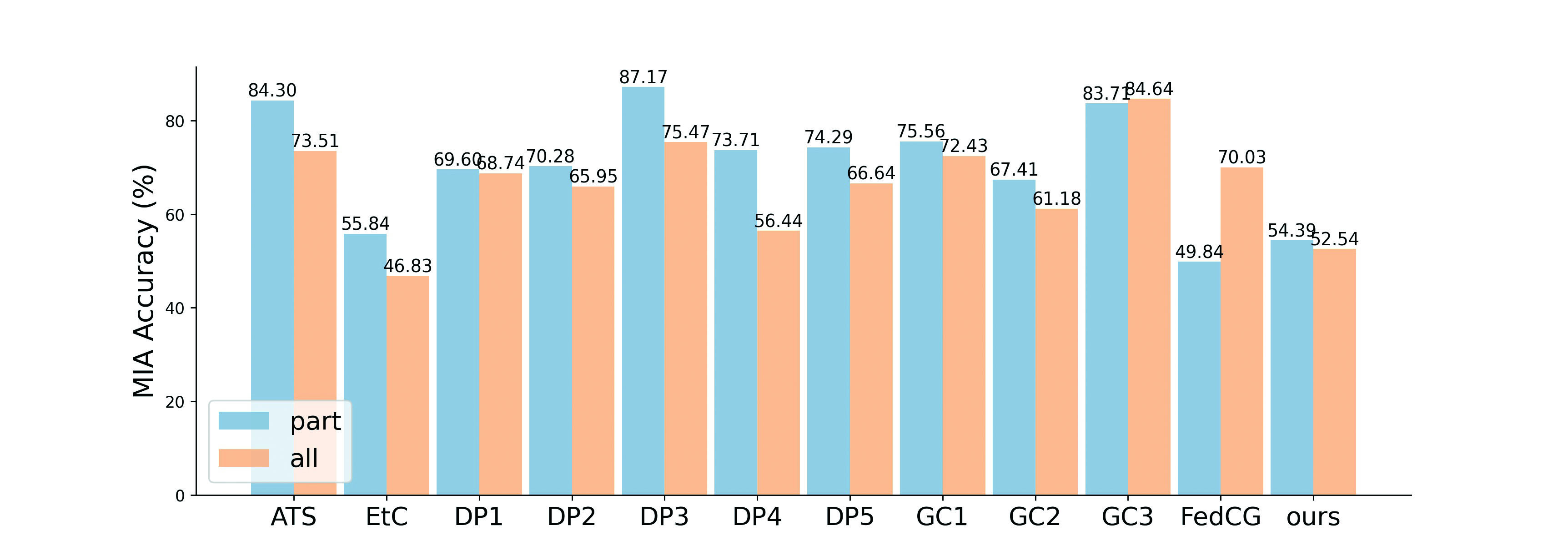}
    \caption{The MIA accuracy of different data partitions under different defense policies in the CIFAR10 dataset. FedCG protects from attacks by uploading partial parameters to the server.}
\label{fig:MIA_CIFAR10}
\end{figure*}

\begin{figure*}[t]
    \centering
    \includegraphics[width=1.0\linewidth]{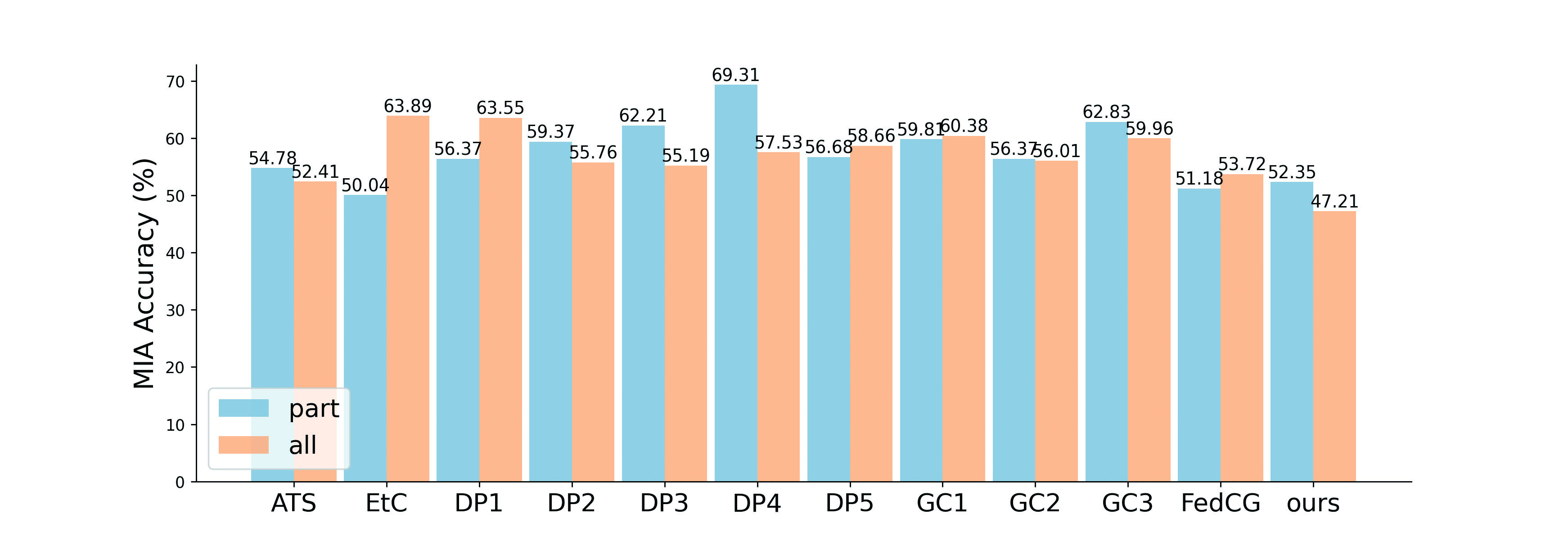}
    \caption{The MIA accuracy of different data partition under different defense policies in the SVHN dataset. The SVHN dataset is composed of numbers with the same general characteristics, which are more difficult to attack. Our method is still able to achieve a low success rate among all defense methods.}
\label{fig:MIA_SVHN}
\end{figure*}

The full experiment results of this attack are shown in Figure \ref{fig:MIA_MNIST}, Figure \ref{fig:MIA_FMNIST}, Figure \ref{fig:MIA_CIFAR10}, and Figure \ref{fig:MIA_SVHN}.
More accuracy details are shown in Table \ref{tab:MIA_MNISTdetail}, Table \ref{tab:MIA_FMNISTdetail}, Table \ref{tab:MIA_CIFAR10detail}, and Table \ref{tab:MIA_SVHNdetail}.
Due to the structure diversity of the original text, we use ResNet18 as the classification model in \emph{ATS} and \emph{EtC} policy.
In other methods, we follow the original settings (basically LeNet).

\subsection{Image Reconstruction Attack}

\begin{figure*}[t]
    \centering
    \includegraphics[width=1.0\linewidth]{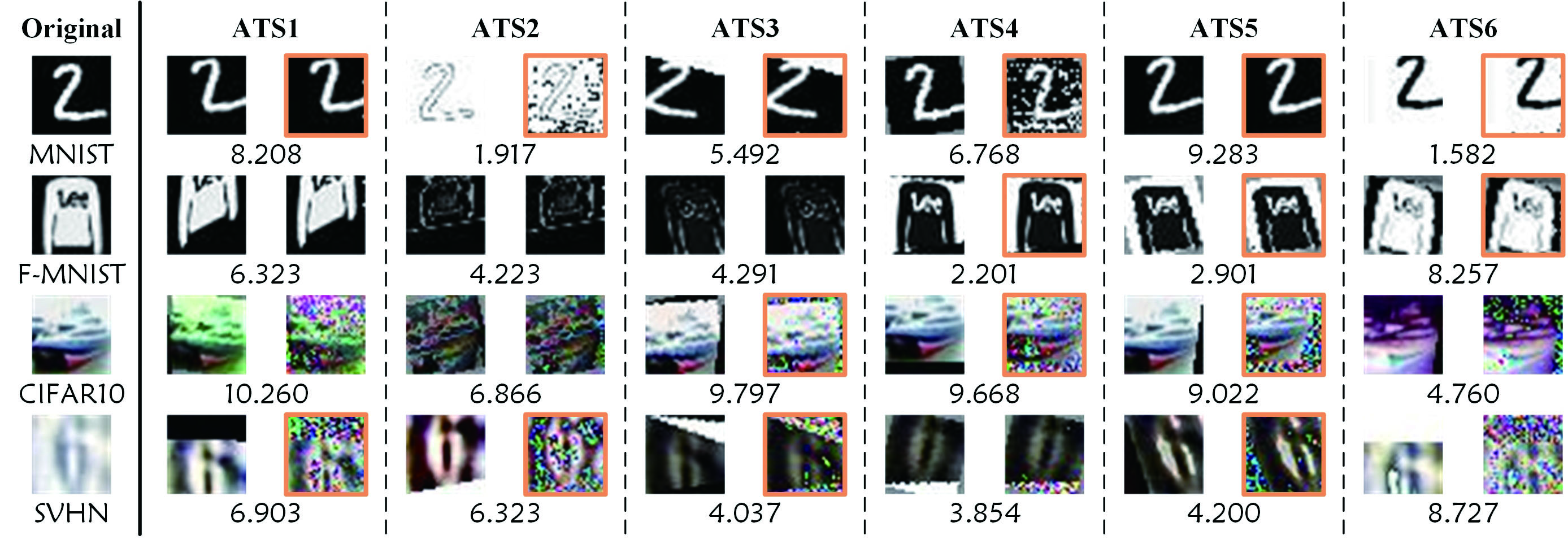}
\caption{The reconstruction attack in ATS policies with different data transformation methods. We randomly apply the strategy six times, each of which goes through three different data augmentation methods. The left figure in each policy is the image after transformation. The right figure is the recovered image. We use PSNR(dB) to display the quantitative result between the original image and the recovered image. Most of the images can be restored to the extent that they have been transformed by ATS. Not all ATS can resist attacks, related to the random combination of data augmentations.}
\label{fig:RS_ATS}
\end{figure*}

\begin{figure*}[t]
    \centering
    \includegraphics[width=1.0\linewidth]{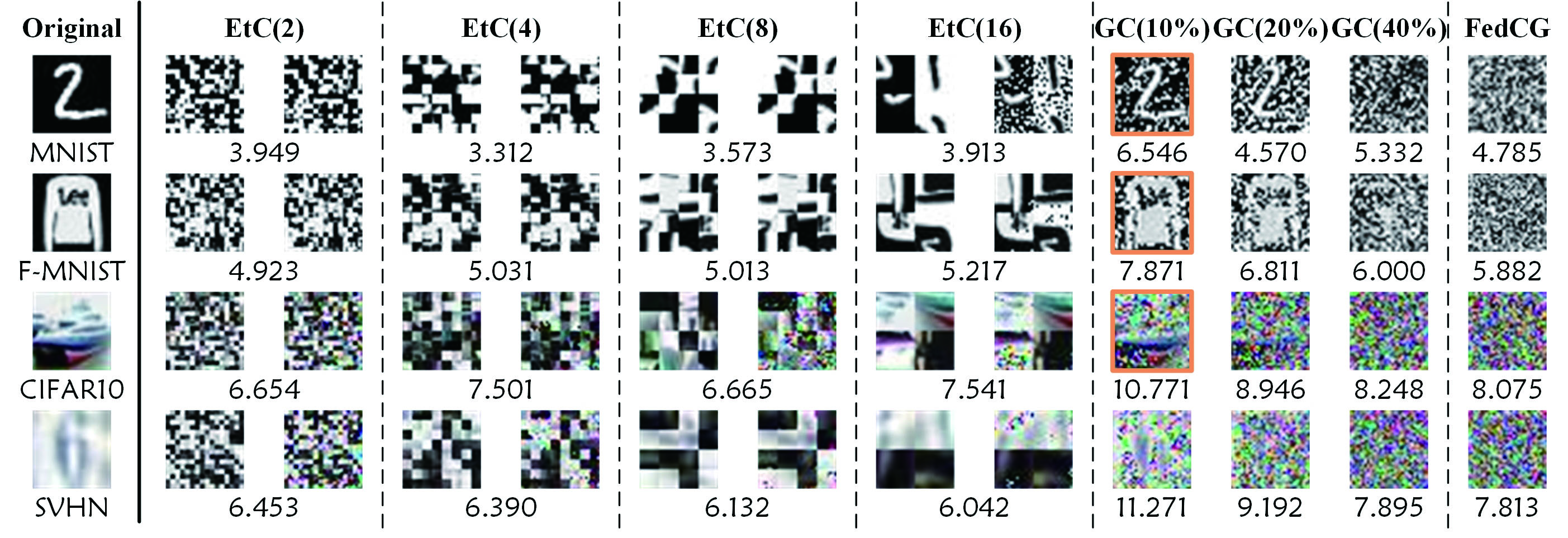}
\caption{The reconstruction attack in EtC policies (different block sizes), GC policies (varying degrees of gradient compression), and FedCG policy. The numbers in brackets denote the size of each image block and the compression percentage. In the EtC policy, even if the attacker can recover images (right image in each policy) from gradients, only the encrypted image (left image in each policy) can be recovered, which is theoretically safe since the attacker cannot decrypt encrypted images without the encryption algorithm. We resize the size of images of MNIST and F-MNIST datasets to 32 $\times$ 32 in EtC. In GC policy, the outlines of the original image are basically visible at 10\% compression, and cannot be recovered at 40\% compression. However, the higher the compression, the worse the model utility. FedCG can protect users from reconstruction attacks.}
\label{fig:RS_EtCGC}
\end{figure*}

\begin{figure*}[t]
    \centering
    \includegraphics[width=1.0\linewidth]{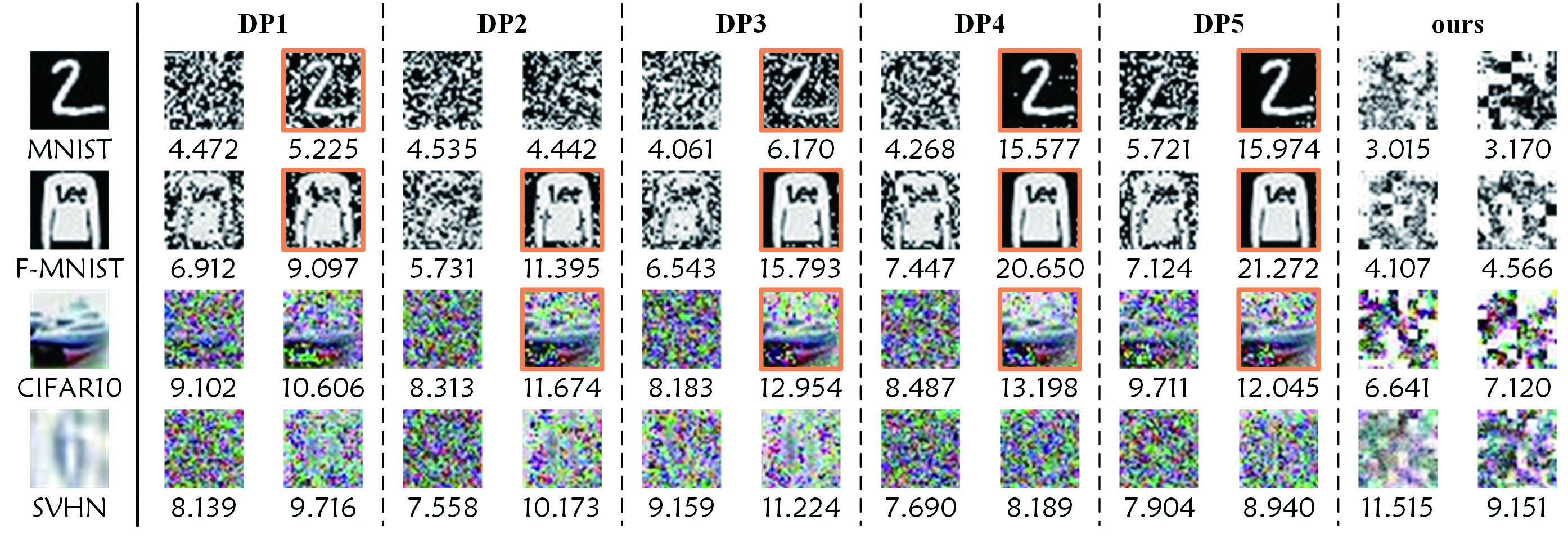}
\caption{The reconstruction attack in DP policies (different data ratios) and our proposal. In each DP policy, the left figure represents the ``part'' situation (the data size in sensitivity is one batch) whereas the right one denotes the ``all'' situation (the data size in sensitivity is all local data). It is largely impossible to recover the original image in the ``part'' scenario, while a rough outline can be restored in the ``all'' scenario, however, it is still essentially unable to be recovered in the SVHN dataset. In our method, the left figure shows the recovered image by the input of dummy data, and the right one is generated from data with the same distribution as the target dataset. Our scheme can protect privacy from reconstruction attacks. Using original images instead of noise as inputs further prevents attackers from reconstructing images. This is one of the reasons why we use CycleGAN.}
\label{fig:RS_DPours}
\end{figure*}

We implement reconstruction attacks on the LeNet network for 200 epochs.
We change the label to a one-hot label and carry out this attack on one specific image with the L-BFGS optimizer (learning rate is 1). 
The full experimental results are shown in Figure \ref{fig:RS_ATS}, Figure \ref{fig:RS_EtCGC}, and Figure \ref{fig:RS_DPours}.

\begin{table*}[t]
\centering
\begin{tabular}{c|c|ccc|ccc|ccc}
\hline
\textbf{Policy} &                               & \multicolumn{3}{c|}{\textbf{32}} & \multicolumn{3}{c|}{\textbf{64}} & \multicolumn{3}{c}{\textbf{128}} \\ \cline{3-11} 
                &                               & Sigmoid   & ReLU    & LReLU      & Sigmoid   & ReLU    & LReLU      & Sigmoid   & ReLU    & LReLU           \\ \hline
{\textbf{DP}}   &\textless{}5,10\textgreater{}  & 100       & 90.63   & 87.50      & 100       & 90.63   &  92.19     & 100       & 97.66   & 96.88           \\
                &\textless{}10,10\textgreater{} & 100       & 87.50   & 90.63      & 100       & 90.63   &  93.75     & 100       & 95.31   & 96.88           \\
                &\textless{}20,10\textgreater{} & 100       & 87.50   & 87.50      & 100       & 92.19   &  92.19     & 100       & 96.09   & 96.88           \\
                &\textless{}20,5\textgreater{}  & 100       & 87.50   & 87.50      & 100       & 92.19   &  93.75     & 100       & 96.09   & 96.09           \\
                &\textless{}20,20\textgreater{} & 100       & 90.63   & 87.50      & 100       & 90.63   &  92.19     & 100       & 96.88   & 96.88           \\ \hline
\textbf{GC}     &p=10\%                         & 100       & 93.75   & 93.75      & 100       & 89.06   & 93.75      & 100       & 92.19   & 90.63            \\
                &p=20\%                         & 100       & 93.75   & 93.75      & 100       & 92.19   & 89.06      & 100       & 90.63   & 92.19            \\
                &p=40\%                         & 100       & 93.75   & 90.63      & 100       & 90.63   & 93.75      & 100       & 91.41   & 91.41            \\ \hline
\end{tabular}
\caption{Label inference attack accuracy (\%) of different batch sizes and activation functions in the protection of DP (CIFAR10) and GC (SVHN) policy. DP\textless{}$\epsilon$, C\textgreater{}, where C refers to the clipping bound and the privacy budget is $\epsilon$. The full setting of DP can been seen in the source code. p denotes the different degrees of gradient compression. With three different batch sizes, there is no significant change in the attack success rate.} 
\label{LIA_DP}
\end{table*}

\begin{table*}[t]
\centering
\begin{tabular}{c|ccc|ccc|ccc|ccc}
\hline
{\textbf{FedCG}} & \multicolumn{3}{c|}{\textbf{MNIST}} & \multicolumn{3}{c|}{\textbf{F-MNIST}} & \multicolumn{3}{c|}{\textbf{CIFAR10}}  & \multicolumn{3}{c}{\textbf{SVHN}}\\ \cline{2-13} 
                             & 32  & 64   & 128   & 32  & 64   & 128   & 32  & 64   & 128  & 32  & 64   & 128 \\ \hline
Acc(\%)              & 100       & 100       & 100       & 100       & 100       & 100     & 100       & 100       & 100       & 100       & 100       & 100            \\ \hline
\end{tabular}
\caption{Label inference attack accuracy of different batch sizes and datasets in the protection of FedCG policy.} 
\label{LIA_FedCG}
\end{table*}

\begin{table*}[t]
\centering
\begin{tabular}{cc|ccc||ccc}
\hline
\multicolumn{2}{c|}{{\textbf{Policy}}}                              & \multicolumn{3}{c||}{\textbf{part}}                       & \multicolumn{3}{c}{\textbf{all}}                         \\ \cline{3-8} 
\multicolumn{2}{c|}{}                                                              & \textbf{Precision$\uparrow$} & \textbf{Recall$\uparrow$} & \textbf{F1-score$\uparrow$} & \textbf{Precision$\uparrow$} & \textbf{Recall$\uparrow$} & \textbf{F1-score$\uparrow$} \\ \hline
\multicolumn{2}{c|}{\textbf{ATS}}                                                  & 0.50(0.51)         & 0.70(0.31)      & 0.59(0.39)        & 0.10(0.90)         & 0.80(0.20)      & 0.18(0.33)        \\ \hline
\multicolumn{2}{c|}{\textbf{EtC}}                                                  & 0.50(0.49)         & 0.58(0.41)      & 0.53(0.45)        & 0.10(0.90)         & 0.39(0.60)      & 0.16(0.72)        \\ \hline
\multicolumn{1}{c|}{{\textbf{DP}}} & \textless{}5,10\textgreater{}  & 0.54(0.54)         & 0.54(0.54)      & 0.54(0.54)        & 0.10(0.90)         & 0.45(0.57)      & 0.17(0.70)        \\
\multicolumn{1}{c|}{}                             & \textless{}10,10\textgreater{} & 0.51(0.51)         & 0.61(0.41)      & 0.56(0.45)        & 0.10(0.90)         & 0.63(0.38)      & 0.18(0.54)        \\
\multicolumn{1}{c|}{}                             & \textless{}20,10\textgreater{} & 0.51(0.52)         & 0.58(0.45)      & 0.54(0.48)        & 0.10(0.90)         & 0.64(0.37)      & 0.18(0.53)        \\
\multicolumn{1}{c|}{}                             & \textless{}20,5\textgreater{}  & 0.51(0.52)         & 0.61(0.42)      & 0.56(0.46)        & 0.10(0.90)         & 0.63(0.39)      & 0.18(0.54)        \\
\multicolumn{1}{c|}{}                             & \textless{}20,20\textgreater{} & 0.52(0.51)         & 0.47(0.56)      & 0.49(0.53)        & 0.10(0.90)         & 0.71(0.29)      & 0.18(0.44)        \\ \hline
\multicolumn{1}{l|}{{\textbf{GC}}} & p=10\%                         & 0.52(0.52)         & 0.49(0.56)      & 0.51(0.54)        & 0.11(0.91)         & 0.48(0.56)      & 0.18(0.70)        \\
\multicolumn{1}{l|}{}                             & p=20\%                         & 0.52(0.53)         & 0.59(0.47)      & 0.55(0.50)        & 0.10(0.90)         & 0.54(0.48)      & 0.17(0.63)        \\
\multicolumn{1}{l|}{}                             & p=40\%                         & 0.53(0.53)         & 0.52(0.54)      & 0.52(0.53)        & 0.10(0.90)         & 0.55(0.47)      & 0.17(0.62)        \\ \hline
\multicolumn{2}{c|}{\textbf{FedCG}}                                                & 0.51(0.51)         & 0.60(0.41)      & 0.55(0.46)        & 0.10(0.90)         & 0.79(0.21)      & 0.18(0.34)        \\ \hline
\multicolumn{2}{c|}{\textbf{ours}}                                                 & 0.46(0.42)         & 0.59(0.30)      & 0.51(0.35)        & 0.09(0.87)         & 0.61(0.30)      & 0.15(0.45)        \\ \hline
\end{tabular}
\caption{Membership inference attack details in the MNIST dataset. The numbers in brackets represent data from non-member, while those outside parentheses represent member data. $\uparrow$ represents a better attack effect and worse privacy-preserving ability. In the part case, the random guess has a 50\% success rate because the two users have the same amount of data. And in the all scenario, victims only constitute 10\% of the data volume of all users. The dataset is simple, membership inference is hard to attack.} 
\label{tab:MIA_MNISTdetail}
\end{table*}

\begin{table*}[t]
\centering
\begin{tabular}{cc|ccc||ccc}
\hline
\multicolumn{2}{c|}{{\textbf{Policy}}}                              & \multicolumn{3}{c||}{\textbf{part}}                       & \multicolumn{3}{c}{\textbf{all}}                         \\ \cline{3-8} 
\multicolumn{2}{c|}{}                                                              & \textbf{Precision$\uparrow$} & \textbf{Recall$\uparrow$} & \textbf{F1-score$\uparrow$} & \textbf{Precision$\uparrow$} & \textbf{Recall$\uparrow$} & \textbf{F1-score$\uparrow$} \\ \hline
\multicolumn{2}{c|}{\textbf{ATS}}                                                  & 0.55(0.56)         & 0.60(0.50)      & 0.57(0.53)        & 0.11(0.91)         & 0.58(0.48)      & 0.18(0.63)        \\ \hline
\multicolumn{2}{c|}{\textbf{EtC}}                                                  & 0.52(0.52)         & 0.53(0.50)      & 0.53(0.51)        & 0.11(0.91)         & 0.49(0.55)      & 0.18(0.68)        \\ \hline
\multicolumn{1}{c|}{{\textbf{DP}}} & \textless{}5,10\textgreater{}  & 0.58(0.58)         & 0.57(0.59)      & 0.58(0.58)        & 0.13(0.92)         & 0.54(0.60)      & 0.21(0.73)        \\
\multicolumn{1}{c|}{}                             & \textless{}10,10\textgreater{} & 0.61(0.58)         & 0.52(0.67)      & 0.56(0.62)        & 0.14(0.92)         & 0.50(0.67)      & 0.22(0.78)        \\
\multicolumn{1}{c|}{}                             & \textless{}20,10\textgreater{} & 0.61(0.58)         & 0.51(0.67)      & 0.55(0.62)        & 0.11(0.91)         & 0.64(0.42)      & 0.19(0.58)        \\
\multicolumn{1}{c|}{}                             & \textless{}20,5\textgreater{}  & 0.55(0.55)         & 0.58(0.53)      & 0.56(0.54)        & 0.12(0.91)         & 0.50(0.58)      & 0.19(0.71)        \\
\multicolumn{1}{c|}{}                             & \textless{}20,20\textgreater{} & 0.58(0.56)         & 0.48(0.66)      & 0.52(0.60)        & 0.11(0.92)         & 0.64(0.43)      & 0.19(0.59)        \\ \hline
\multicolumn{1}{c|}{{\textbf{GC}}} & p=10\%                         & 0.64(0.61)         & 0.58(0.67)      & 0.61(0.64)        & 0.16(0.93)         & 0.56(0.67)      & 0.25(0.78)        \\
\multicolumn{1}{c|}{}                             & p=20\%                         & 0.54(0.57)         & 0.69(0.41)      & 0.61(0.48)        & 0.13(0.92)         & 0.51(0.61)      & 0.20(0.73)        \\
\multicolumn{1}{c|}{}                             & p=40\%                         & 0.61(0.67)         & 0.75(0.52)      & 0.67(0.58)        & 0.13(0.92)         & 0.53(0.61)      & 0.21(0.73)        \\ \hline
\multicolumn{2}{c|}{\textbf{FedCG}}                                                & 0.50(0.49)                   &  0.60(0.30)   &  0.54(0.44)     & 0.10(0.90)          & 0.40(0.60)     &  0.16(0.72)         \\ \hline
\multicolumn{2}{c|}{\textbf{ours}}                                                 & 0.50(0.50)         & 0.40(0.60)      & 0.44(0.55)        & 0.12(0.92)         & 0.59(0.50)      & 0.19(0.65)        \\ \hline
\end{tabular}
\caption{Membership inference attack details in the Fashion-MNIST dataset. The dataset is more complex than the MNIST dataset, so the precision is improved.} 
\label{tab:MIA_FMNISTdetail}
\end{table*}

\begin{table*}[t]
\centering
\begin{tabular}{cc|ccc||ccc}
\hline
\multicolumn{2}{c|}{{\textbf{Policy}}}                              & \multicolumn{3}{c||}{\textbf{part}}                       & \multicolumn{3}{c}{\textbf{all}}                         \\ \cline{3-8} 
\multicolumn{2}{c|}{}                                                              & \textbf{Precision$\uparrow$} & \textbf{Recall$\uparrow$} & \textbf{F1-score$\uparrow$} & \textbf{Precision$\uparrow$} & \textbf{Recall$\uparrow$} & \textbf{F1-score$\uparrow$} \\ \hline
\multicolumn{2}{c|}{\textbf{ATS}}                                                  & 0.83(0.86)         & 0.87(0.82)      & 0.85(0.84)        & 0.24(0.97)         & 0.76(0.73)      & 0.37(0.83)        \\ \hline
\multicolumn{2}{c|}{\textbf{EtC}}                                                  & 0.59(0.54)         & 0.38(0.73)      & 0.46(0.62)        & 0.10(0.90)         & 0.56(0.46)      & 0.17(0.61)        \\ \hline
\multicolumn{1}{c|}{{\textbf{DP}}} & \textless{}5,10\textgreater{}  & 0.70(0.69)         & 0.69(0.71)      & 0.69(0.70)        & 0.19(0.95)         & 0.66(0.69)      & 0.30(0.80)        \\
\multicolumn{1}{c|}{}                             & \textless{}10,10\textgreater{} & 0.71(0.69)         & 0.68(0.73)      & 0.70(0.71)        & 0.18(0.95)         & 0.68(0.66)      & 0.28(0.78)        \\
\multicolumn{1}{c|}{}                             & \textless{}20,10\textgreater{} & 0.88(0.87)         & 0.87(0.88)      & 0.87(0.87)        & 0.25(0.96)         & 0.72(0.76)      & 0.37(0.85)        \\
\multicolumn{1}{c|}{}                             & \textless{}20,5\textgreater{}  & 0.74(0.74)         & 0.74(0.74)      & 0.74(0.74)        & 0.14(0.93)         & 0.64(0.56)      & 0.23(0.70)        \\
\multicolumn{1}{c|}{}                             & \textless{}20,20\textgreater{} & 0.75(0.73)         & 0.72(0.76)      & 0.74(0.75)        & 0.19(0.95)         & 0.71(0.66)      & 0.30(0.78)        \\ \hline
\multicolumn{1}{l|}{{\textbf{GC}}} & p=10\%                         & 0.78(0.74)         & 0.71(0.80)      & 0.75(0.77)        & 0.23(0.96)         & 0.75(0.72)      & 0.35(0.82)        \\
\multicolumn{1}{l|}{}                             & p=20\%                         & 0.69(0.66)         & 0.64(0.71)      & 0.66(0.68)        & 0.16(0.94)         & 0.65(0.61)      & 0.25(0.74)        \\
\multicolumn{1}{l|}{}                             & p=40\%                         & 0.85(0.82)         & 0.81(0.86)      & 0.83(0.84)        & 0.38(0.98)         & 0.82(0.85)      & 0.52(0.91)        \\ \hline
\multicolumn{2}{c|}{\textbf{FedCG}}                                                & 0.50(0.50)         & 0.60(0.40)      & 0.54(0.44)        & 0.11(0.90)         & 0.27(0.75)      & 0.15(0.82)        \\ \hline
\multicolumn{2}{c|}{\textbf{ours}}                                                 & 0.57(0.53)         & 0.35(0.74)      & 0.43(0.62)        & 0.09(0.89)         & 0.40(0.54)      & 0.14(0.67)        \\ \hline
\end{tabular}
\caption{Membership inference attack details in the CIFAR10 dataset. Because the image complexity is higher than several other datasets, the member inference results are best.} 
\label{tab:MIA_CIFAR10detail}
\end{table*}

\begin{table*}[t]
\centering
\begin{tabular}{cc|ccc||ccc}
\hline
\multicolumn{2}{c|}{{\textbf{Policy}}}                              & \multicolumn{3}{c||}{\textbf{part}}                       & \multicolumn{3}{c}{\textbf{all}}                         \\ \cline{3-8} 
\multicolumn{2}{c|}{}                                                              & \textbf{Precision$\uparrow$} & \textbf{Recall$\uparrow$} & \textbf{F1-score$\uparrow$} & \textbf{Precision$\uparrow$} & \textbf{Recall$\uparrow$} & \textbf{F1-score$\uparrow$} \\ \hline
\multicolumn{2}{c|}{\textbf{ATS}}                                                  & 0.54(0.55)         & 0.61(0.48)      & 0.58(0.52)        & 0.12(0.92)         & 0.61(0.51)      & 0.21(0.66)        \\ \hline
\multicolumn{2}{c|}{\textbf{EtC}}                                                  & 0.50(0.50)         & 0.47(0.54)      & 0.48(0.52)        & 0.10(0.90)         & 0.32(0.67)      & 0.15(0.77)        \\ \hline
\multicolumn{1}{c|}{{\textbf{DP}}} & \textless{}5,10\textgreater{}  & 0.57(0.56)         & 0.55(0.58)      & 0.56(0.57)        & 0.13(0.91)         & 0.45(0.66)      & 0.20(0.76)        \\
\multicolumn{1}{c|}{}                             & \textless{}10,10\textgreater{} & 0.59(0.60)         & 0.64(0.55)      & 0.61(0.57)        & 0.11(0.91)         & 0.47(0.57)      & 0.18(0.70)        \\
\multicolumn{1}{c|}{}                             & \textless{}20,10\textgreater{} & 0.60(0.65)         & 0.71(0.53)      & 0.65(0.59)        & 0.12(0.92)         & 0.56(0.55)      & 0.20(0.69)        \\
\multicolumn{1}{c|}{}                             & \textless{}20,5\textgreater{}  & 0.73(0.67)         & 0.61(0.77)      & 0.67(0.71)        & 0.12(0.91)         & 0.50(0.58)      & 0.19(0.71)        \\
\multicolumn{1}{c|}{}                             & \textless{}20,20\textgreater{} & 0.57(0.56)         & 0.55(0.59)      & 0.56(0.58)        & 0.11(0.91)         & 0.46(0.60)      & 0.18(0.72)        \\ \hline
\multicolumn{1}{l|}{{\textbf{GC}}} & p=10\%                         & 0.60(0.60)         & 0.61(0.59)      & 0.60(0.59)        & 0.13(0.92)         & 0.54(0.61)      & 0.21(0.74)        \\
\multicolumn{1}{l|}{}                             & p=20\%                         & 0.57(0.56)         & 0.54(0.58)      & 0.55(0.57)        & 0.12(0.91)         & 0.51(0.57)      & 0.19(0.70)        \\
\multicolumn{1}{l|}{}                             & p=40\%                         & 0.62(0.63)         & 0.65(0.60)      & 0.64(0.62)        & 0.16(0.95)         & 0.70(0.58)      & 0.25(0.72)        \\ \hline
\multicolumn{2}{c|}{\textbf{FedCG}}                                                & 0.51(0.51)         & 0.59(0.43)      & 0.55(0.47)        & 0.11(0.91)         & 0.49(0.54)      & 0.17(0.68)        \\ \hline
\multicolumn{2}{c|}{\textbf{ours}}                                                 & 0.53(0.52)         & 0.41(0.64)      & 0.46(0.57)        & 0.09(0.89)         & 0.49(0.47)      & 0.16(0.62)        \\ \hline
\end{tabular}
\caption{Membership inference attack details in the SVHN dataset.} 
\label{tab:MIA_SVHNdetail}
\end{table*}

\end{document}